\documentclass[10pt,twocolumn,letterpaper]{article}
\pdfoutput=1
\usepackage{cvpr}
\usepackage{times}
\usepackage{epsfig}
\usepackage{graphicx}
\usepackage{amsmath}
\usepackage{amssymb}

\usepackage[pagebackref=true,breaklinks=true,letterpaper=true,colorlinks,bookmarks=false]{hyperref}

\cvprfinalcopy

\def\cvprPaperID{****}

\ifcvprfinal\fi

\usepackage[utf8]{inputenc}
\usepackage[T1]{fontenc}
\usepackage{hyperref}
\usepackage{url}
\usepackage{booktabs}
\usepackage{amsfonts}

\usepackage{amsmath}
\usepackage{amsfonts}
\usepackage{amssymb}
\usepackage{amsthm}
\usepackage{algorithm}
\usepackage{placeins}
\usepackage{algpseudocode}
\usepackage{subcaption}
\usepackage{csquotes}

\newtheorem{theorem}{Theorem}
\theoremstyle{definition}

\graphicspath{{./figures}}

\begin{document}

\title{Orthogonal Wasserstein GANs}

\author{Jan M{\"u}ller\\
University of Bonn\\
{\tt\small mueller.jan.u@gmail.com}
\and
Reinhard Klein\\
University of Bonn\\
{\tt\small rk@cs.uni-bonn.de}
\and
Michael Weinmann\\
University of Bonn\\
{\tt\small mw@cs.uni-bonn.de}
}

\maketitle

\begin{abstract}
Wasserstein-GANs have been introduced to address the deficiencies of generative adversarial networks (GANs) regarding the problems of vanishing gradients and mode collapse during the training, leading to improved convergence behaviour and improved image quality.
However, Wasserstein-GANs require the discriminator to be Lipschitz continuous.
In current state-of-the-art Wasserstein-GANs this constraint is enforced via gradient norm regularization.
In this paper, we demonstrate that this regularization does not encourage a broad distribution of spectral-values in the discriminator weights, hence resulting in less fidelity in the learned distribution.
We therefore investigate the possibility of substituting this Lipschitz constraint with an orthogonality constraint on the weight matrices.
We compare three different weight orthogonalization techniques with regards to their convergence properties, their ability to ensure the Lipschitz condition and the achieved quality of the learned distribution. 
In addition, we provide a comparison to Wasserstein-GANs trained with current state-of-the-art methods, where we demonstrate the potential of solely using orthogonality-based regularization.
In this context, we propose an improved training procedure for Wasserstein-GANs which utilizes orthogonalization to further increase its generalization capability.
Finally, we provide a novel metric to evaluate the generalization capabilities of the discriminators of different Wasserstein-GANs.  
\end{abstract}

\section{Introduction}
Generative modelling has gained attention due to its improvements for numerous applications including semi-supervised learning, image and 3D modeling, data completion or super-resolution.
Inspired by game theory, generative adversarial networks (GANs) are based on the competition of two players -- represented in terms of respective generator and discriminator networks – where the generator tries to generate samples so that the discriminator cannot distinguish whether they are real or generated samples.
In the original definition, the objective to be minimized is given by the Jensen-Shannon divergence \cite{goodfellow2014}, which is a symmetric extension of the Kullback-Leibler divergence and measures the overlap between two distributions.
However, GANs in their original formulation face several problems such as lacking stability during training, which includes vanishing gradients, mode collapse, as well as a non-converging loss for both generator and discriminator.
The Kantorovich duality \cite{villani2008optimal} allows the Kullback-Leibler divergence to be replaced by the Wasserstein distance, which mitigates the convergence problem due to the preservation of gradient information, guarantees differentiability of the objective function, and less susceptibility to mode-collapse, partially by enabling the discriminator to differentiate between overlapping manifolds \cite{arjovsky2017}.
This requires enforcing a Lipschitz constraint (introduced by the Kantorovich duality) on the discriminator as the unconstrained problem would result in exploding gradients. This can be achieved by clipping the weights to lie within a compact interval \cite{arjovsky2017}.
Other methods soften this constraint by a regularization with the gradient norm to improve the framework's robustness with regard to different architectures and the quality of generated samples \cite{salimans2016improved, petzka2018on}.
In this paper, we will demonstrate among other things that this regularization does not encourage a broad distribution of spectral-values in the discriminator weights.
A narrow distribution of singular values results in a model which is unable to capture all details of the distribution~\cite{miyato2018spectral}.
Approaches which have been valuable in the context of standard GANs such as spectral normalization (SN)~\cite{miyato2018spectral} can only improve a WGAN's stability when used in addition to a gradient penalty.
We found in initial experiments that WGANs regularized only with SN did not converge, which is consistent with Miyato’s comment~\cite{miyato-comment}.
SN forces a network to learn a Lipschitz continuous function by bounding the 2-norm of the weights.
The discriminator of a WGAN has a gradient norm of 1 almost everywhere. Therefore, according to Theorem 1 and 2 by Anil \etal~\cite{anil2018sorting} orthogonality is necessary.
However, these improvements were achieved at the cost of of a higher computational burden due to the additional regularization term that has to be also considered during backpropagation, which dramatically increased training time of WGANs when compared to the original GAN framework.
Initially WGAN discriminators were trained until convergence (or at least multiple steps) before the generator was updated.
While this problem has been addressed by the two times-scale update rule \cite{heusel2017gans} which allows to reduce the number of discriminator updates per generator update and influenced the training of more recent architectural methods to reduce the computational complexity such as the progressive insertion of layers \cite{karras2017progressive}, the additional costs of computing the regularization remain.
Furthermore we demonstrate that the two times-scale update rule leads to a reduced ability in capturing the modes of a distribution. 
In this paper, we direct our attention on increasing the fidelity of learned distribution by investigating the possibility of substituting the Lipschitz constraint required by Wasserstein-GANs with an orthogonality constraint on the weight matrices during training. The major contributions of this work are:
\begin{itemize}
  \item We investigate the possibility to replace the Lipschitz constraint with an orthogonality constraint on the weights, where we compare three weight orthogonalization methods regarding their convergence properties, their ability to ensure the Lipschitz condition and the achieved quality of the learned distribution.
  \item We introduce a new metric to compare Wasserstein-GAN discriminators based on their approximated Wasserstein distance in order to compare their fitness, i.e. the generalization capabilities of discriminators.
  \item We demonstrate the benefits of using weight orthogonalization during the training of Wasserstein-GANs to enforce its Lipschitz constraint and increase its generalization capability.
\end{itemize}

\section{Background}

As we focus on the use of orthogonality constraints to enforce the Lipschitz constraint in the WGAN setting, we first provide a general overview regarding orthogonality regularization for CNNs.
This is followed by a review of the Wasserstein objective function for GANs~\cite{arjovsky2017}, a discussion of improvements for Wasserstein GANs and a survey of standard evaluation measures applied for comparing the performances of GANs.
\subsection{Orthogonality regularization for CNNs}
The training of deep convolutional neural networks (CNNs) is complicated by a multitude of phenomena such as vanishing/exploding gradients or shifting feature statistics~\cite{ioffe2015batch}.
Besides solutions such as parameter initialization, residual connections, and normalization of
internal activations~\cite{ioffe2015batch}, much attention has been paid to regularization.
In particular, structural regularization such as the energy-preserving orthogonality regularization has been explored to stabilize the optimization and increase its efficiency~\cite{Rodriguez:2017,Desjardins2015NaturalNN}.
Further investigations~\cite{Jia:2017,Harandi:2016,Ozay:2016,Xie:2017,Huang:2018} proposed the use of specialized orthogonality regularizations or constraints for various tasks such as using Stiefel manifold-based hard orthogonality constraints of weights~\cite{Harandi:2016,Ozay:2016,Huang:2018} during optimization or using a singular value bounding (SVB)~\cite{Jia:2017}, \ie enforcing the singular values of weight matrices to be close to one based on a pre-specified threshold.
Recent work~\cite{Xie:2017} additionally investigated soft orthonormal regularization by penalizing deviations of each weight matrix' Gram matrix to the identity matrix in the Frobenius sense.
The benefits of such soft orthonormal regularization are its differentiability and its reduced computational burden due to not relying on singular value decomposition.
However, Frobenius norm-based orthogonality regularization represents only a rough approximation and may be inaccurate especially for dense matrices.
Other work focused on penalizing the spectral norm of weight matrices in CNNs~\cite{Yoshida:2017}.
A further generalization of soft orthogonality regularization to non-square weight matrices and consistent performance gains for different network architectures has been achieved by Basal et al.~\cite{Bansal:2018} that introduced double soft orthogonality regularization, mutual coherence regularization and spectral restricted isometry property regularization.
While impressive results have been achieved (especially with spectral restricted isometry property regularization), the extension of enforcing orthogonality in the training of GANs has been left as future work.

\subsection{Wasserstein objective function for GANs}
Let $(\mathcal{X}, d)$ be a compact metric space with $\sigma$-algebra $\Sigma$.
We denote the set of probability distributions over $\mathcal{X}$ with $\text{prob}(\mathcal{X})$ and the distribution of real data as $\mathbb{P}_r \in \text{prob}(\mathcal{X})$.
Furthermore, $Z$ is a random variable over a space $\mathcal{Z}$ and we assume an a-priori probability density $p_\mathcal{Z}$ for $Z$.

The distance between two distributions $\mathbb{P}, \mathbb{Q} \in \text{prob}(\mathcal{X})$ can be measured by the $1$-Wasserstein distance
\begin{equation*}
  \label{eq:wdistance}
  W_1(\mathbb{P}, \mathbb{Q}) = \inf_{\pi \in \Pi(\mathbb{P}, \mathbb{Q})} \int_{\mathcal{X}} d(x,y) d\pi(x,y)
\end{equation*}
where $d$ is the metric on $\mathcal{X}$, $\Pi(\mathbb{P}, \mathbb{Q})$ denotes the set of all joint distributions over $\mathcal{X}^2$ whose marginals are $\mathbb{P}$ and $\mathbb{Q}$.
Its dual presentation given by the Kantorovich duality \cite{villani2008optimal} is the following optimization problem over the set of real valued $1$-Lipschitz continuous functions:
\begin{equation}
  \label{eq:wdistance_dual}
    W_1(\mathbb{P}, \mathbb{Q}) = \sup_{ \lVert f \rVert_{\text{lip}} \leq 1} \int_{\mathcal{X}} f dp - \int_{\mathcal{X}} f dq
\end{equation}
The Wasserstein-GAN can now be modelled by two parametrized functions $f_{\omega} : \mathcal{X} \to \mathbb{R}$ and $g_\theta: \mathcal{Z} \to \mathcal{X}$ where $f_\omega$ denotes the critic and $g_\theta$ the generator.
The generator (with its objective to optimize $g_\theta$ and produce samples that cannot be distinguished from real samples) and the discriminator (with its objective to approximate the dual potential $f$ with a parametrized function $f_\omega$ ) compete in the minimax game
\begin{equation*}
  \min_{\theta} \max_{\omega} \mathbb{E}_{x \sim \mathbb{P}_r}[f_\omega(x)] - \mathbb{E}_{z \sim p(z)}[f_\omega(g_\theta(z))].
\end{equation*}
When implemented, this minimax game is relaxed and the critic is trained until convergence and the Lipschitz constraint is enforced either via clipping the weights in a compact space~\cite{arjovsky2017} or regularizing the critics objective with an estimated gradient norm~\cite{gulrajani2017improved}.
\subsection{Improving Wasserstein GANs}
Enforcing the Lipschitz constraint on the Wasserstein-GAN's discriminator is crucial to ensure the models convergence.
Numerous normalization procedures have been demonstrated to increase a networks adversarial robustness by limiting its Lipschitz constant \cite{NIPS2018_7515}.
Most prominent in the literature on GANs are instance/batch-normalization. 
Other techniques such as weight-normalization \cite{salimans2016weight} have been found to be limiting when compared to spectral normalization \cite{miyato2018spectral}.
However, we have found that weight and spectral normalization do not ensure a successful training, although they limit the discriminator's Lipschitz constant.
Only additional regularization with the gradient norm lead a successful training of a Wasserstein-GAN and we discuss its theoretical problems in Section \ref{ssec:enforcing}.
In its original formulation, the Lipschitz constraint is enforced by clipping the weights so that they are contained in a compact interval \cite{arjovsky2017}.
Based on the theoretical insight that an optimal discriminator has the gradient norm $1$ almost everywhere \cite{gulrajani2017improved}, further improvements have been made by enforcing this constraint with regularization \cite{gulrajani2017improved,petzka2018on}.
These improvements increase the stability of the model and quality of the generated images but require additional computation during training.
The application of a two time-scale update rule for GANs (WGAN-TTUR) allows to reduce the number of discriminator updates per generator update and further enhances the convergence properties and the sample quality \cite{heusel2017gans}.
We demonstrate that WGAN-TTUR decreases the ability to represent all modes of the real distribution, and introduce a trainings procedure to mitigate this problem by allowing the network to increase its capacity.
Further relevant investigations, which focus on improvements to the networks architectures, include the progressive insertion of layers~\cite{karras2017progressive} and the use of large conditional GANs~\cite{brock2018large}.
The progressive insertion of layers by fade-in~\cite{karras2017progressive} further increases the computational efficiency and quality on image datasets by architectural means. 
In contrast, training large-scale conditional GANs~\cite{brock2018large} has been approached based on a hinge version of the original GAN objective and a regularization by conditioning the GAN according to the large annotated JFT-300m dataset to mitigate mode collapse.
However, these improvements are specific to large-scale GANs and not related to improving WGANs that mitigate mode collapse based on the Wasserstein objective function.
In this paper, we will discuss the suitability of soft orthogonality enforcing techniques as well as hard constraint and discuss why constraint on the norm are not sufficient to enforce the Lipschitz constraint.
\subsection{Measures for evaluating GANs}
When applied to image datasets the current state-of-the-art approach in automatically evaluating the image quality are Inception-Score \cite{salimans2016improved} and the Frechet-Inception-Distance (FID) \cite{heusel2017gans}.
However, the score computed by both methods becomes better if the network overfits.
Methods to directly evaluate overfitting or mode-collapse in GANs \cite{srivastava2017veegan,arora2017gans,santurkar2018classification} either require human supervision or knowledge about the modes or label distribution of a dataset.
An estimate of the Wasserstein distance between local image features \cite{karras2017progressive} denoted as sliced Wasserstein distance (SWD) provides a value that indicates the difficulty to distinguish real from generated images, however, its high computational complexity makes this approach less feasible.
In contrast, we propose a novel and easy to compute WGAN evaluation metric that scores the models' generalization capabilities based on the estimated Wasserstein distance.

\section{What can we gain from orthogonality regularization?}
Wasserstein-GANs~\cite{arjovsky2017} have been introduced to mitigate the major problems of standard GANs~\cite{goodfellow2014} regarding their unstable training, vanishing gradients, strange convergence behaviour and mode-collapse.
However, enforcing the Lipschitz constraint introduced by the Kantorovich duality in Equation \ref{eq:wdistance_dual} is necessary as an unconstrained maximization problem would diverge and the discriminator would provide no meaningful gradient to the generator.
In this section, we demonstrate drawbacks of previous methods to enforce the Lipschitz constraint and elucidate how Wasserstein-GANs can benefit from an orthogonal weight constraint.
\subsection{Problems of regularization based on the gradient norm} \label{ssec:enforcing}
\begin{figure}[!h]
  \centering
  \begin{subfigure}{0.325\linewidth}
    \includegraphics[width=\linewidth]{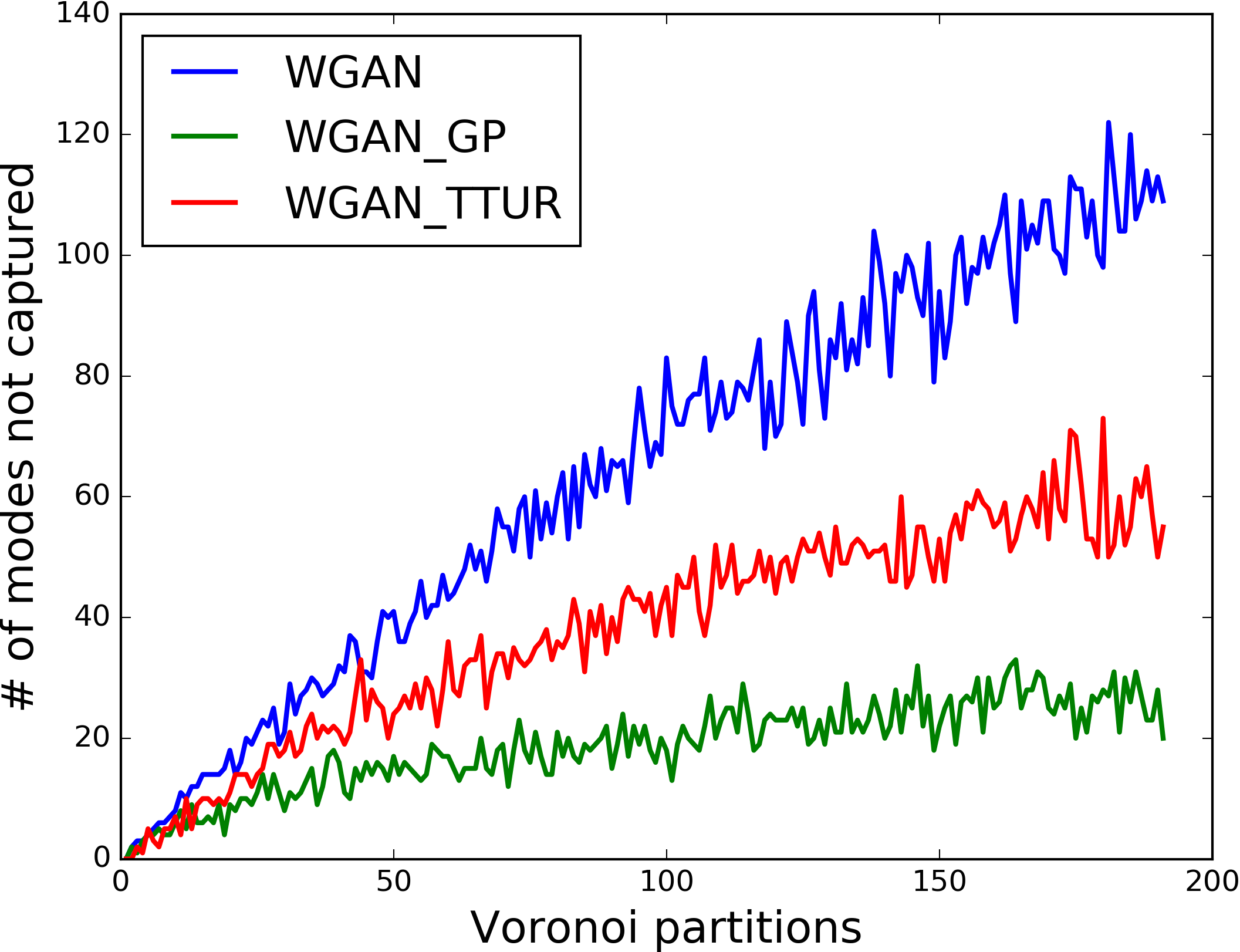} \\
    \caption{
    Number of missing modes in CelebA}
    \label{fig:disadvantages1}
  \end{subfigure}
  \begin{subfigure}{0.325\linewidth}
    \includegraphics[width=\linewidth]{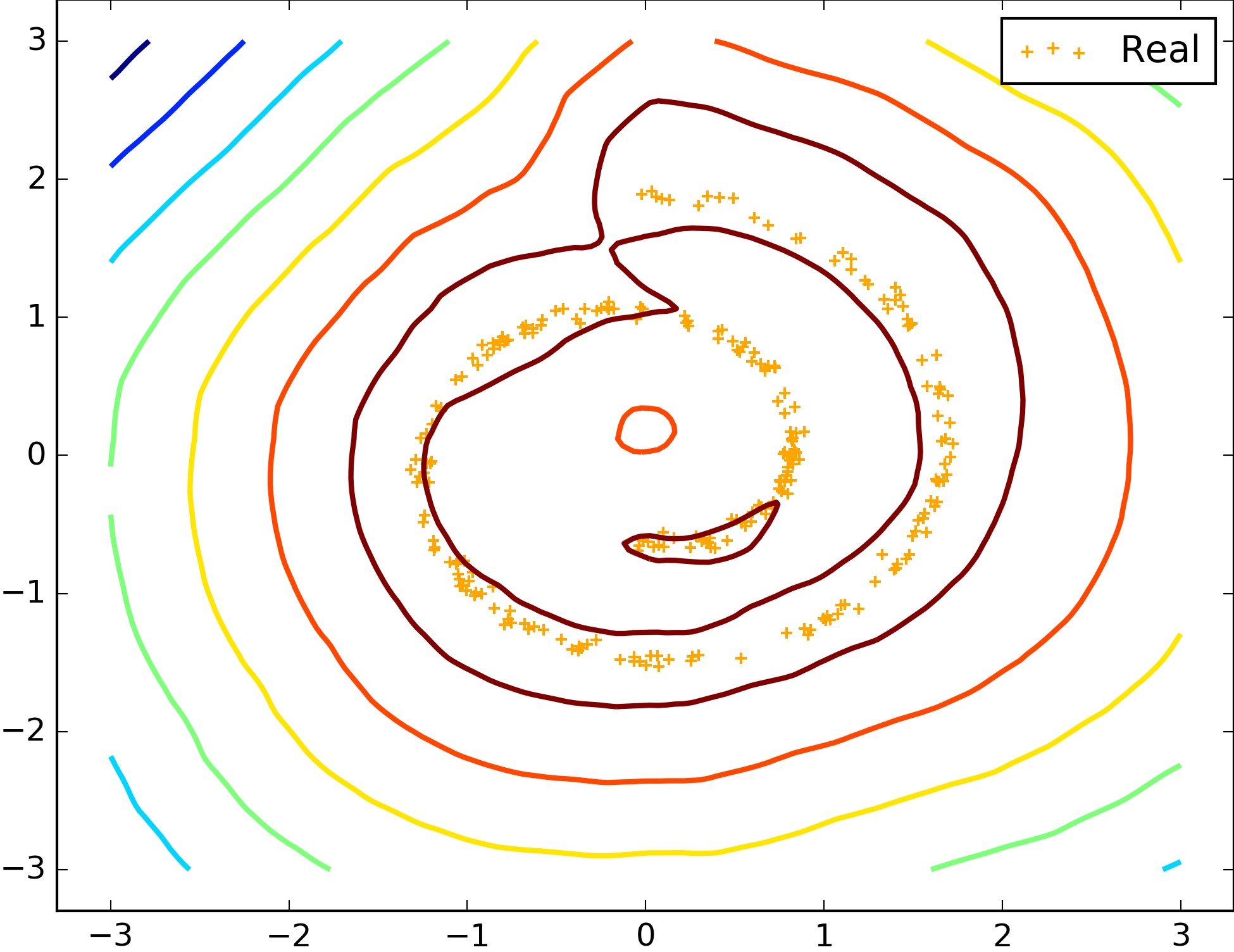} \\
    \caption{WGAN-GP}
    \label{fig:disadvantages2}
  \end{subfigure}
  \begin{subfigure}{0.325\linewidth}
    \includegraphics[width=\linewidth]{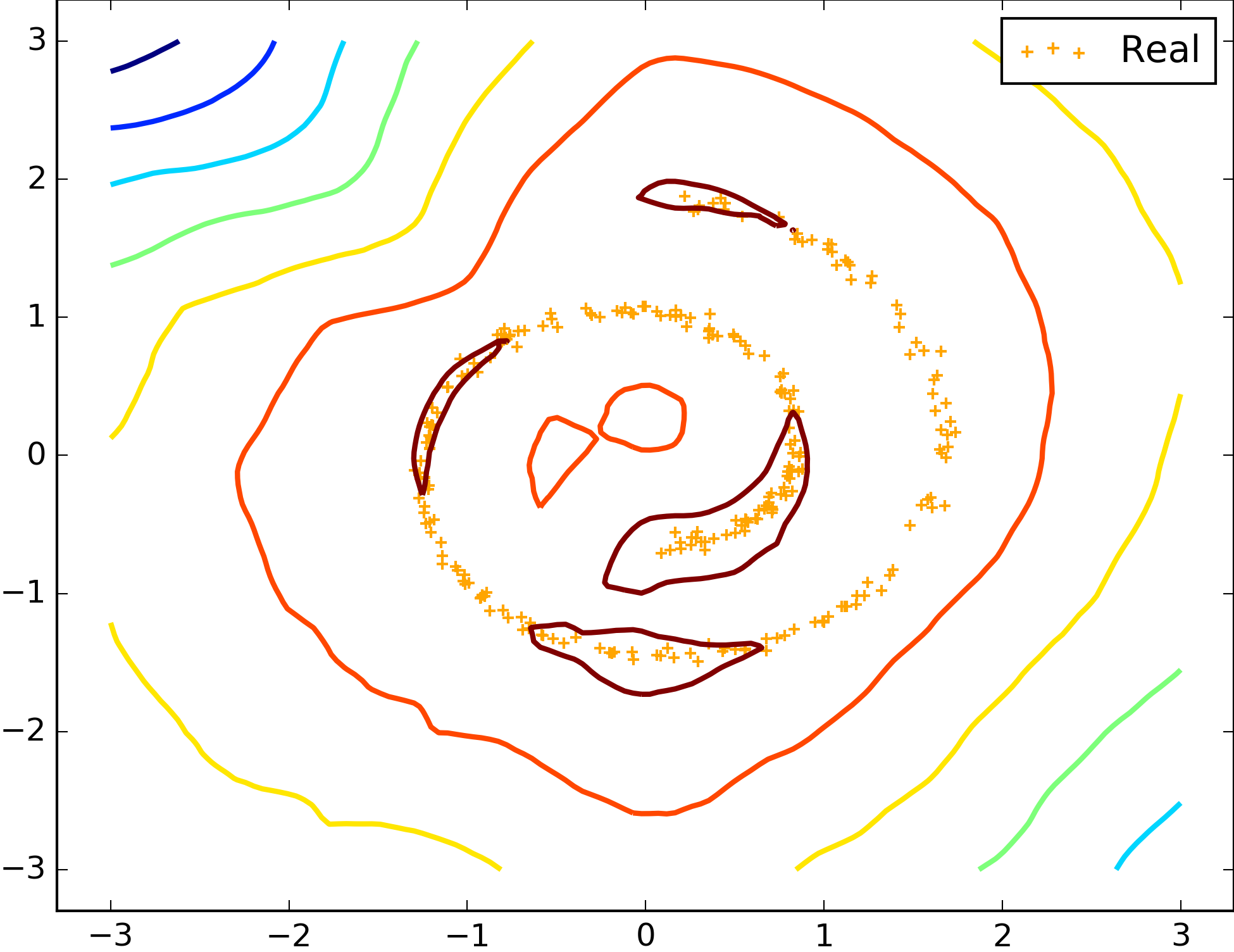} \\
    \caption{WGAN-TTUR}
    \label{fig:disadvantages3}
  \end{subfigure}
	\caption{(a) Number of approximated modes which are statistically significantly less preserved in the generated distribution when compared to its test dataset. (b) Visualization of the discriminator's ability to capture a synthetic distribution as introduced in~\protect\cite{gulrajani2017improved}. When trained with $n_{\text{critic}} = 5$ discriminator updates in between each generator update and the same learning rates for both networks (as used in WGAN-gp), the discriminator's approximation of the dual potential clearly outlines the target distribution. (c) Reducing $n_{\text{critic}}$ to $1$ and applying the TTUR results in a discriminator that does not capture the distribution in the same quality, even though both models have been trained with the same computational budget.}
  \label{fig:disadvantages}
\end{figure}

Stochastic gradient descent does not directly allow for conditional optimization, and therefore additional techniques have been established to enforce the Lipschitz constraint for a neural network which approximate the dual potential $f$.
Methods such as clipping the weights to lie within a compact interval \cite{arjovsky2017} or enforcing $L_2$-constraints do not achieve state-of-the-art results due to the fact that these constraint allow the discriminator to collapse to a linear function \cite{anil2018sorting}.
Recent state-of-the-art methods which aim to minimize a Wasserstein loss \cite{karras2017progressive,adler2018banach} have adopted regularization to enforce the Lipschitz constraint.
The discriminator is regularized with its gradient norm 
\begin{equation}
  \label{eq:gp}
  \mathbb{E}_{x \sim \hat{\mathbb{P}}} \left[ \left( \lVert \nabla_{\hat{x}} f_\omega(\hat{x}) \rVert_2 - 1 \right)^2 \right]
\end{equation}
where $\hat{\mathbb{P}}$ is based on interpolated samples between the generated and target distributions \cite{gulrajani2017improved} to mitigate vanishing/exploding gradients.
Such a regularization increases the computational capacity needed during training by $30\%$ and scales (almost) linearly with the number of layers as demonstrated in the supplemental. 
The improved stability offered by this regularization \cite{gulrajani2017improved} allows to reduce the number of discriminator updates between each generator update to $1$, and instead use a two time-scale update rule (TTUR) \cite{heusel2017gans} to avoid losses in image quality.
In this TTUR, the generator and discriminator are trained in an alternating scheme with different learning rates, allowing the use of a higher learning rate for the discriminator to reduce the trainings time.
However, as demonstrated in Figure \ref{fig:disadvantages}, a Wasserstein-GAN trained according to the two time-scale update rule has a reduced ability to capture the modes of the target distribution.
Furthermore this problem does not only occur for synthetic distributions.
We utilize the method introduced by \cite{richardson2018gans} to evaluate the mode collapse of a Wasserstein-GAN regularized with the gradient penalty from Equation \ref{eq:gp} (WGAN-GP)~\cite{gulrajani2017improved} and a Wasserstein-GAN using the same regularization but trained according to the TTUR on the benchmark dataset CelebA.
The modes are approximated by computing a Voronoi partition. As the true number of modes is unknown for the distribution that is assumed to underlie the datasets, we tested with a range between $1$ and $100$ Voronoi partitions.
A statistical analysis reveals that the number of modes is significantly less well-represented on both the CIFAR-10 dataset~\cite{krizhevsky2009learning} and the CelebA dataset~\cite{Liu:2015:DLF} for WGAN-TTUR in comparison to WGAN-GP as demonstrated in Figure~\ref{fig:disadvantages1}.
Results for other datasets are included in the supplemental.
A relevant question is therefore how the representation can be improved without drastically increasing the training time.
\subsection{Relation between Lipschitz continuity and orthogonality constraints}
Orthogonal weight constraints have been proven to stabilize the training of RNNs \cite{wisdom2016full} and increase the generalization capabilities \cite{Bansal:2018}. 
A quadratic matrix $W \in \mathbb{R}^{n \times n}$ is orthogonal if and only if $W^T W = W W^T = I$.
For simplicity of notation, we call a non-quadratic matrix $W \in \mathbb{R}^{n \times m}$ orthogonal if the matrix has dimensions $n > m$ and orthogonal columns ($W^TW = I$) or dimension $n < m$ and orthogonal rows ($W W^T = I$).
It is well-known that a function $f$ between two metric spaces is Lipschitz continuous if and only if its gradient is bounded.
Let $h(x) = Wx$ be a linear layer with the weight matrix $W$ and inputs $x$, then this implies that $h$ is Lipschitz continuous if $\lVert W \rVert$ is bounded.
If we assume that the discriminator $f_\omega$ is a feedforward network built from $m \in \mathbb{N}$ linear layers $(h_i)_{i \in [1:m]}$ and $1$-Lipschitz continuous activation functions $(a_i)_{i \in [1:m]}$, then
\begin{equation}
  \label{eq:gradient}
  \lVert\nabla_x f\omega(x) \rVert \leq \prod_{i=1}^m \lVert W_i\rVert \lVert \nabla a_i(x) \rVert \leq \prod_{i=1}^m \lVert W_i \rVert  
\end{equation}
\noindent which implies that such a discriminator is Lipschitz continuous if the norm of all weight matrices is bounded.
One might assume that limiting the $2$-norm of the weight matrices would be sufficient to guarantee its Lipschitz constant to be at most $1$.
However, upper-bounding the norm of the network's weight matrices to be at most $1$ without any additional constraint only bounds its Lipschitz constant and does not prevent the network from collapsing to a linear function (assuming $1$-Lipschitz and monotonic activations (such as ReLU) are used)~\cite{anil2018sorting}.
This explains the limited performance reported in \cite{gulrajani2017improved} for hard weight constraints and the limited performance when using spectral normalization \cite{miyato2018spectral} to enforce the discriminator's Lipschitz condition.
\begin{theorem}[\cite{anil2018sorting}]
\label{thm:unitary}
If a function $f: \mathbb{R}^n \to \mathbb{R}$ with $\lVert \nabla f(x) \rVert = 1$ almost everywhere is represented by a neural network with weights $W$ that have a $2$-norm of at most $1$, then $W$ can be replaced by an orthogonal matrix $\hat{W}$ without changing the represented function.
\end{theorem}
The sufficient condition to enforce the Lipschitz constraint of a neural network as provided in Theorem \ref{thm:unitary} is to constrain the weight matrices to be orthogonal.

\section{Orthogonal Wasserstein-GAN}
Motivated by the theoretical connection between a network's Lipschitz constant and an orthogonal weight constraint, we discuss three methods to enforce such a constraint as well as their run-times, and analyse their suitability in the context of training a Wasserstein-GAN.
Based on these findings, we propose a new procedure to train a Wasserstein-GAN.

\subsection{Enforcing Lipschitz constraint with orthogonalization}
An intuitive approach to enforce the orthogonality of the weight matrices is to add regularization to the discriminator's objective according to
\begin{equation}
  \label{eq:reg}
  \lambda \lVert W^T W - I \rVert_{F}^2,
\end{equation}
where $W$ is a weight matrix, $I$ represents the identity matrix, and $\lambda$ weights the contribution of the regularization on the overall objective function.
Such or similar regularization methods have gained an increased adoption in deep neural classifier networks \cite{Bansal:2018} due to the relatively low computational overhead required.
For each layer the computational costs are dominated by computing the matrix multiplication, which scales linearly with the number of layers, but needs additional gradient evaluation.
Orthogonal regularization is only a soft constraint and there is no guarantee that this additional condition is fulfilled. 
The set of all orthogonal matrices is a subspace of $\mathbb{R}^{n \times m}$ called Stiefel manifold.
To perform the optimization on this manifold, the weights should move along the geodesic, for which the direction is given by the gradient $\nabla f(x)$.
Solving optimization problems on the Stiefel manifold has been made tractable with Cayley transformations~\cite{wen2013feasible}
\begin{equation}
  \label{eq:retraction}
  Y(\tau) = \left( I + \frac{\tau}{2}A\right)^{-1} \left( I - \frac{\tau}{2}A\right)
\end{equation}
\noindent where $A = (\nabla_W f(x)) W - (\nabla_W f(x)) W$ is a skew-symmetric $n \times n$ matrix and $\tau$ is the remaining variable to be estimated.
This retraction reduces the optimization problem to the following $m$-dimensional search problem.
For each weight matrix $W_i$ of the network, we now have to find a $\tau_i \in \mathbb{R}$ such that if we set the new weights to be $W_i \gets Y(\tau_i) W_i$ they minimize equation \ref{eq:wdistance_dual}.
However, solving this optimization problem after each generator update does not yield an efficient training for Wasserstein-GANs.
It has been demonstrated that it is sufficient to fix $\tau_i$ to a small value proportional to the learning rate~\cite{wisdom2016full}.
Even though this procedure does not require additional gradient computations, the matrix inversion results in a significantly higher computation burden and higher memory requirements than the regularization according to equation~\ref{eq:reg}.

A more efficient but less accurate orthogonalization algorithm has been introduced by Bj{\"o}rck and Bowie \cite{bjorck1971iterative}. 
For a given weight matrix $W_{0}$ for the step $t=0$, the algorithm iteratively computes the best orthogonal matrix in a least-squares sense by applying
\begin{equation}
  W_{t+1} = W_{t} \left( I + \sum_{i=1}^p (-1)^p { -\frac{1}{2} \choose p } Q_t^p \right)
\end{equation}
\noindent where $t$ is the current iteration and $Q_t = I - W_{t}^T W_{t}$.
Since the algorithm is inherently iterative, it is particularly suitable in the context of neural networks.
We found that the orthogonality and Lipschitz conditions are sufficiently fulfilled by applying one iteration with $p=1$ before each discriminator update.
The asymptotic time complexity is equal to that of regularization but does not require additional gradient computation, which makes it the fastest in an empirical evaluation (see Table~\ref{tab:scores2}).

\subsection{Suitability and comparison of different orthogonality regularizers for WGANs}
We now compare the aforementioned procedures with regard to their suitability in the context of training Wasserstein-GANs.
First, we evaluate the models' adherence to the Lipschitz and orthogonality condition, because a model's convergence behaviour directly depends on its Lipschitz constant. 
The adherence to the orthogonality constraint is quantified by $ \lVert I - W^T W \rVert_2$ for a weight matrix $W$. 
Based on Proposition $1$ in \cite{gulrajani2017improved} we estimate the networks Lipschitz constant with equation \ref{eq:gp}, where the points are drawn from the convex combination of the supports from $\mathbb{P}_r$ and $\mathbb{P}_\theta$.
\begin{figure}
  \centering
  \begin{subfigure}{0.325\linewidth}
    \includegraphics[width=\linewidth]{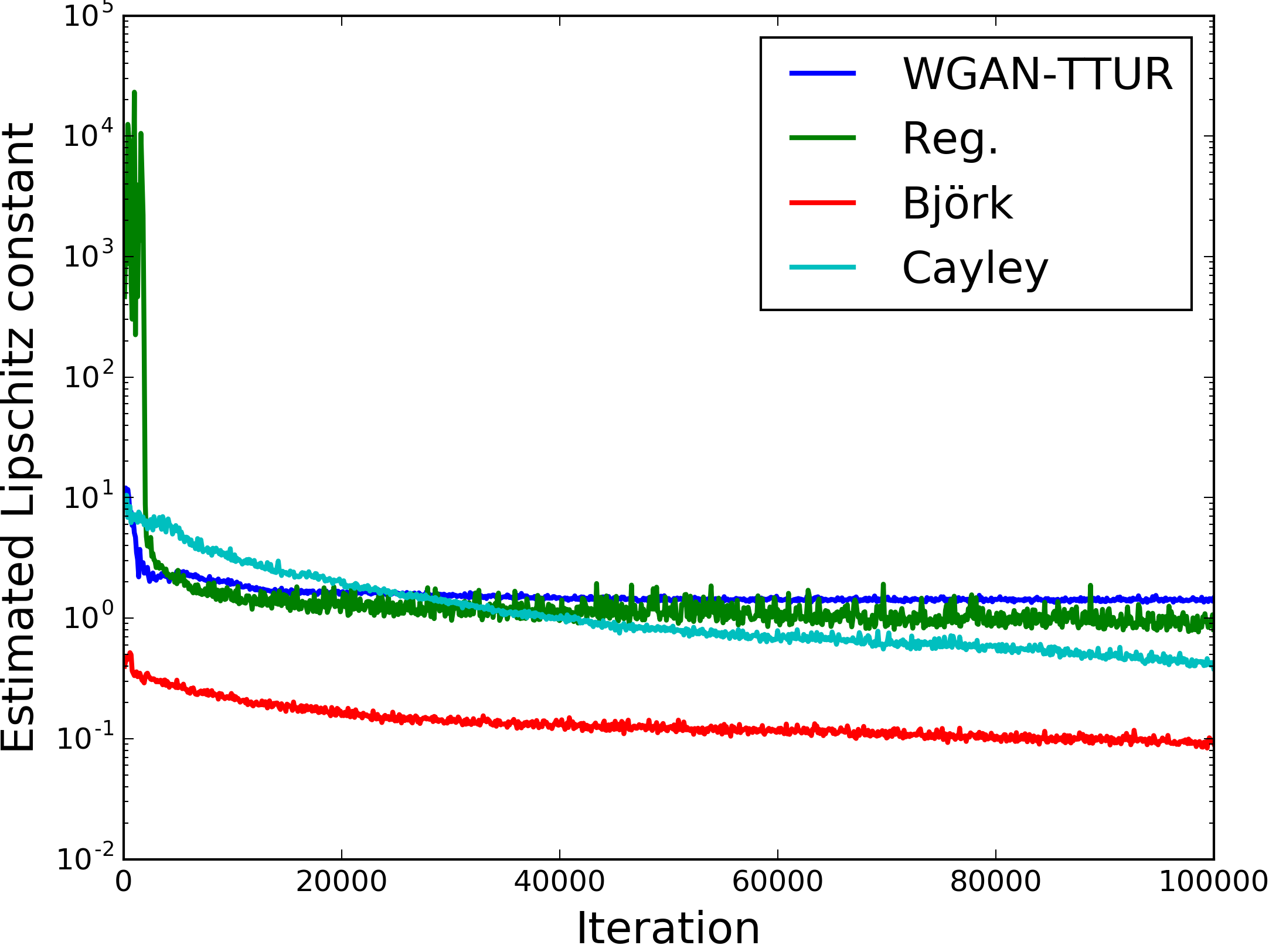} \\
    \caption{estimated Lipschitz constant}
  \end{subfigure}
  \hspace{0.5in}
  \begin{subfigure}{0.325\linewidth}
    \includegraphics[width=\linewidth]{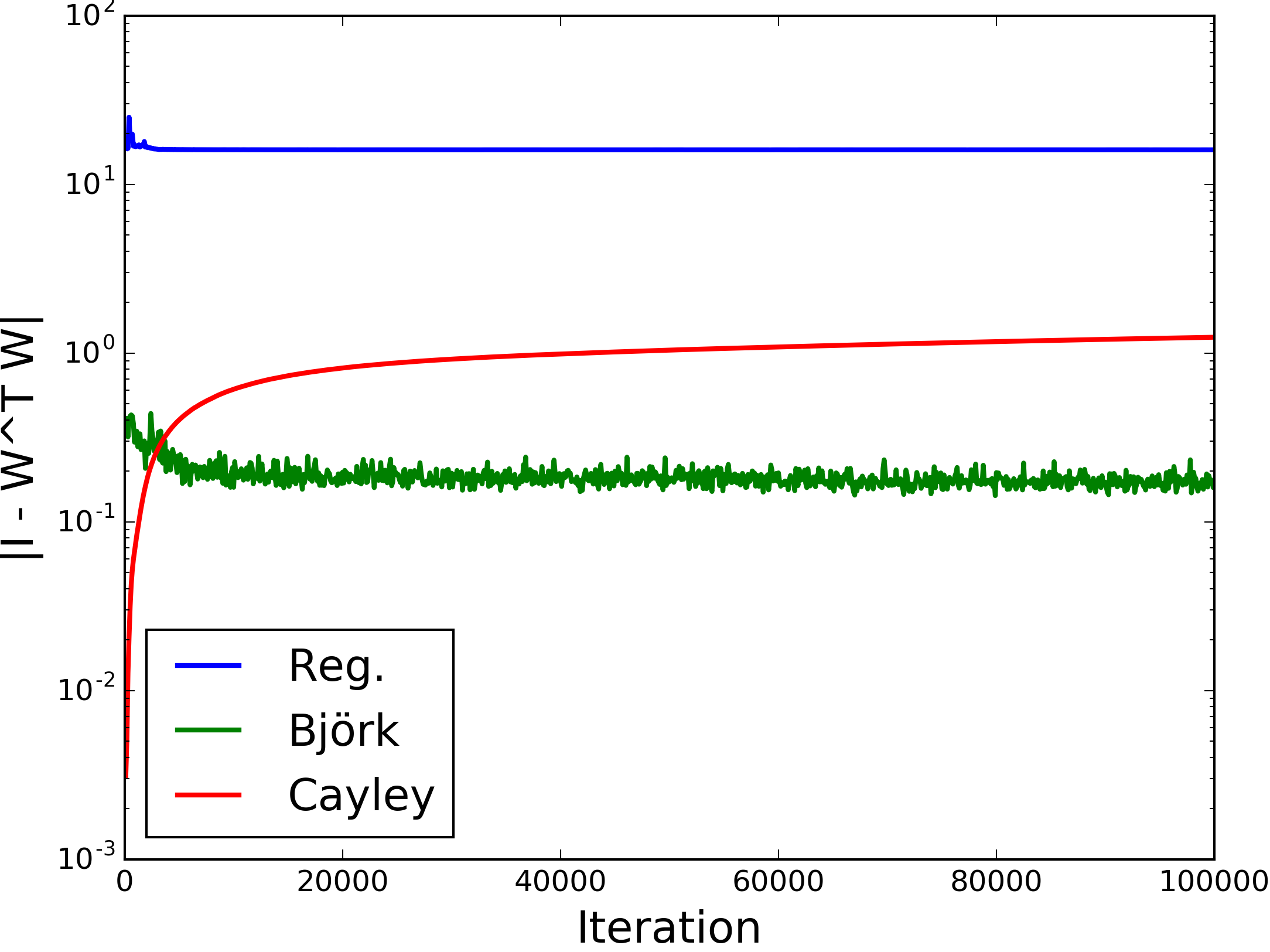} \\
    \caption{Deviation from orthogonal matrix}
  \end{subfigure}
  \caption{Adherence to the Lipschitz and orthogonality constraints. The estimated Lipschitz-constant should be close to or below  one. The deviation from the orthogonal matrix should be close to zero. Note the log scale in both plots.}
  \label{fig:lipschitz}
\end{figure}
We plot the estimated Lipschitz constant and norm for models trained on CIFAR-10 with each of the three methods in Figure \ref{fig:lipschitz}.
We see that all models converge and the Lipschitz constant is bounded in all cases.
However, orthogonal regularization does not ensure orthogonal weight matrices, even for high values of $\lambda$ such as $\lambda = 10$, and we observe a drift with Cayley transformations, which we believe to be a result of numerical inaccuracy.
Iterative orthogonalization enforces both constraints while being significantly faster in comparison to Cayley transformations and comparable in speed to orthogonal regularization as shown in Table \ref{tab:scores2}.

\begin{figure}
  \centering
  \begin{subfigure}{0.325\linewidth}
    \includegraphics[width=\linewidth]{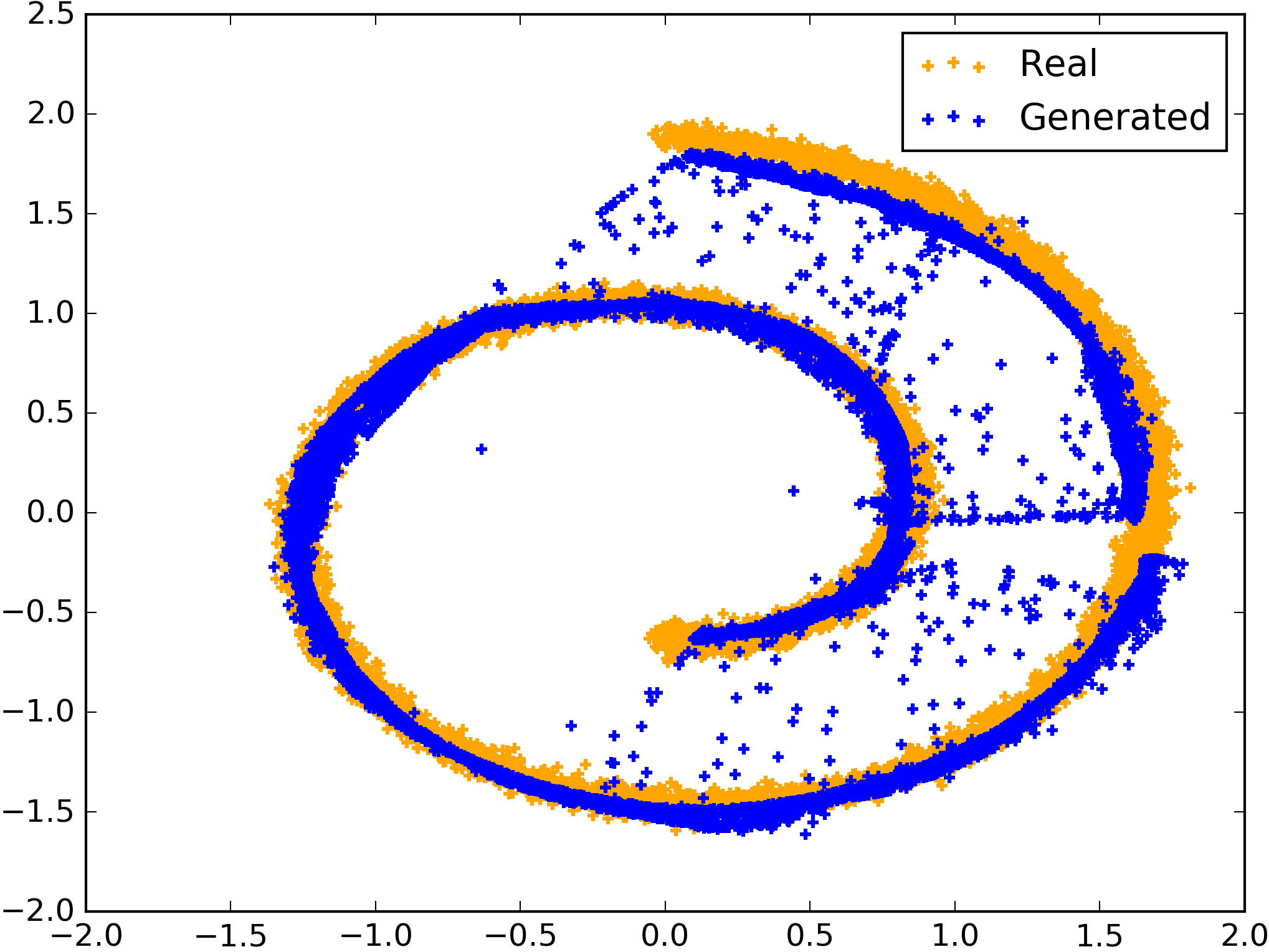} \\
    \caption{Orthogonal regularization.}
  \end{subfigure}
  \begin{subfigure}{0.325\linewidth}
    \includegraphics[width=\linewidth]{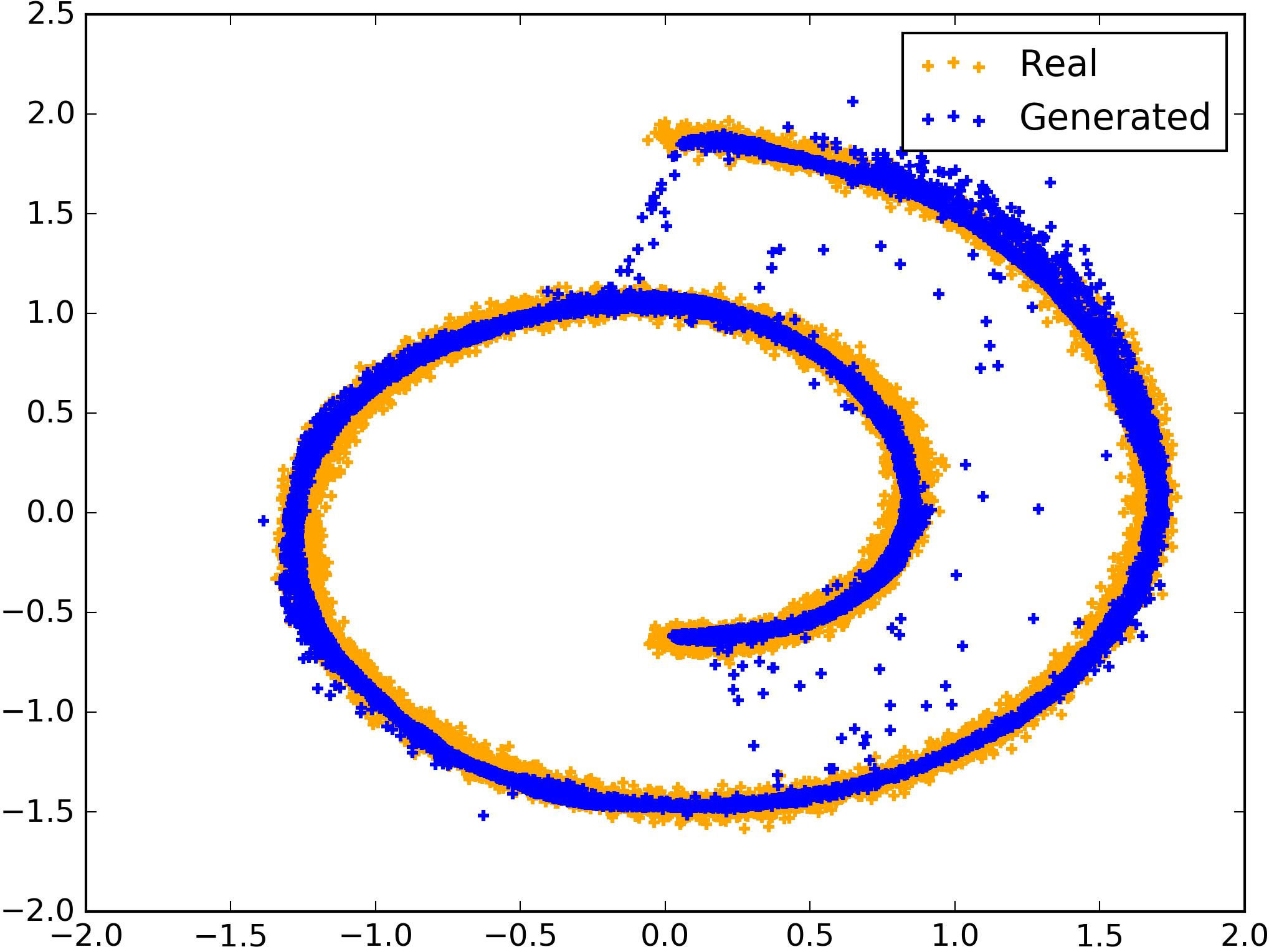} \\
    \caption{Iterative orthogonalization.}
  \end{subfigure}
  \begin{subfigure}{0.325\linewidth}
    \includegraphics[width=\linewidth]{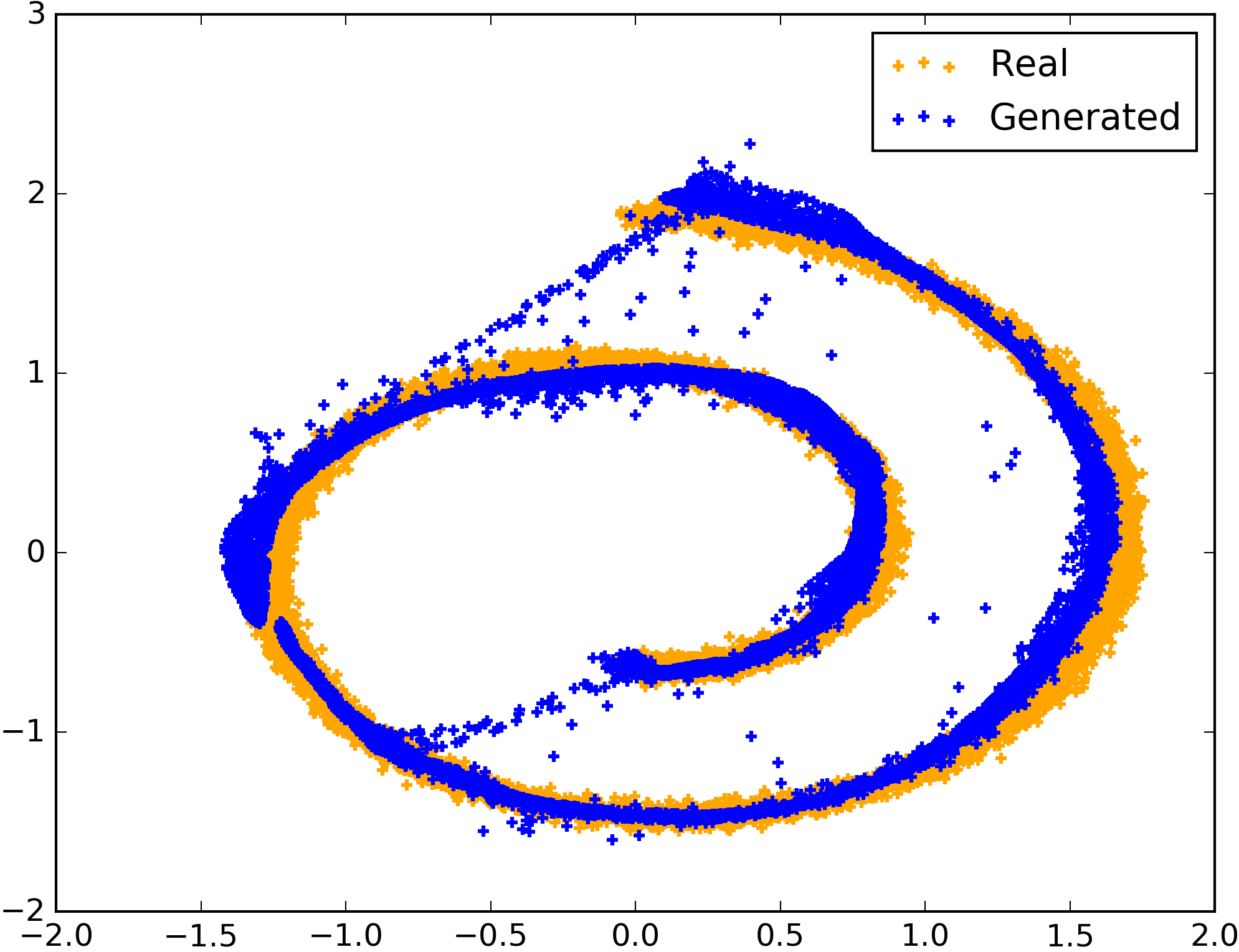} \\
    \caption{Cayley transformation}
  \end{subfigure}
  \caption{Synthetic distribution generated by Wasserstein-GANs trained with orthogonalization for an equal amount of time.}
  \label{fig:synthetic}
\end{figure}

The comparison between the learned synthetic distribution and the real distribution as illustrated in Figure \ref{fig:synthetic} shows that a Wasserstein-GAN trained with iterative orthogonalization captures the target distribution best.
The regularized orthogonalization and the Cayley transformation method both introduce noise, shifts, and distortions in the learned distribution, whereas the iterative orthogonalization method is significantly less affected by these phenomena.
Similar results can be observed in Table \ref{tab:scores2} that quantifies the quality of sampled images from a learned CIFAR-10 representation in terms of both Inception Score (IS) and Fr\'{e}chet Inception Distance (FID).
The Wasserstein-GAN using the iterative orthogonalization has a significantly higher inception score and lower Fr\'{e}chet Inception distance than a Wasserstein-GAN trained using the two methods. 

\subsection{Proposed method}
In the previous section, we demonstrated that the training of a Wasserstein-GAN converges when we only apply iterative orthogonalization in the discriminator.
Note that a solution to equation \ref{eq:wdistance_dual} is only feasible if the discriminator's Lipschitz constant is smaller than 1.
If we compare the estimated Lipschitz constant in Figure \ref{fig:lipschitz}, we observe that a Wasserstein-GAN trained using iterative orthogonalization in the discriminator reaches a feasible solution with fewer iterations than WGAN-TTUR.
However, the resulting scores are not better than WGAN-TTUR's scores as shown in Table \ref{tab:scores2}.
The strict orthogonalization strongly increases the discriminator's robustness against adversarial samples, hinders the discriminator from collapsing to a linear function and shows a faster convergence of its Lipschitz constant.
The normalization of the row and column vectors resulting from orthogonalization leads to less fidelity in the learned distribution \cite{miyato2018spectral}.
In our proposed method, we use the advantages provided by orthogonalization during the beginning of the models training.
As the changes to the generator's output are largest during this initial training phase, we leverage the increased stability provided by iterative orthogonalization during this phase.
We relax this condition for the later training phase and ensure the Lipschitz condition using the one-sided gradient normalization introduced in \cite{petzka2018on}.
A detailed description of our procedure is provided in Algorithm \ref{alg:projection}.
\begin{algorithm}
  \begin{algorithmic}
    \Require discriminator learning rate $\eta_d = 3 \cdot 10^{-4}$, generator learning rate $\eta_g=1 \cdot 10^{-4}$, batchsize $m=64$, $k=\frac{n}{10}$. 
    \For{$i = 1, ..., n$}
    \State $\sigma \gets \text{sigmoid}(i - k)$ 
    \State Sample mini-batch $(x_i)_{i \in [1:m]}$ with $x_i \sim \mathbb{P}_r$.
    \State Sample mini-batch $(z_i)_{i \in [1:m]}$ with $z_i \sim \mathcal{Z}$.
    \State Sample mini-batch $(\hat{x}_i)_{i \in [1:m]}$ with $\hat{x}_i \sim \hat{\mathbb{P}}$.
    \State $g_\omega \gets \nabla_\omega \mathbb{E} \left[ f_\omega(x_i) - f_\omega(g_\theta(z_i))\right] + \lambda \sigma  \mathbb{E} [ \max \{ 0, \lVert \nabla f_\omega(\hat{x}) \rVert_2 - 1 \}^2]$
    \State $\omega \gets w + \eta_\omega \cdot \text{Adam}(\omega, g_\omega)$
    \State $\omega \gets \omega \cdot \left(I + (1-\sigma)\frac{1}{2} \left(I - \omega^T \omega \right) \right)$ 
    \State Sample mini-batch $(z_i)_{i \in [1:m]}$ with $z_i \sim \mathcal{Z}$.
    \State $g_\theta \gets \nabla_\theta \mathbb{E}\left[ f_\omega(g_\theta(z_i)) \right]$
    \State $\theta \gets \theta - \eta_g \cdot \text{Adam}(\theta, g_\theta)$
    \EndFor
    \end{algorithmic}
  \caption{Training procedure of a Wasserstein GAN with orthogonal weights in the discriminator.}
  \label{alg:projection}
\end{algorithm}
Note that in an efficient implementation we can neglect the regularization for the first $k$ steps.
We provide additional information regarding the algorithms extensions regarding CNNs and the used initialization in the supplemental.

\section{Experimental results}
In this section, we first introduce a new metric to compare the generalization capabilities between Wasserstein-GAN discriminators.
Subsequently, we compare our method to both the Wasserstein-GAN regularized with gradient penalty (WGAN-GP)~\cite{gulrajani2017improved} and the Wasserstein-GAN trained according to the two time-scale update rule (WGAN-TTUR)~\cite{heusel2017gans}.
As recommended in \cite{lucic2017gans}, we trained all models with an equal computational budget and architecture.
\subsection{New evaluation metric for the generalization capability of WGANs}
While the Inception Score (IS) \cite{salimans2016improved} and Fr\'{e}chet Inception Distance (FID) \cite{heusel2017gans} are well-established metrics to evaluate the perceived image quality of generated samples and to compare different models with a common architecture, neither of them measures overfitting.
Evaluating overfitting in GANs is non-trivial, because the discriminator can overfit with respect to the real data distribution or the generated samples.
A solution to this problem has been presented in the form of a tournament between different GANs in which the generator/discriminator pairs are compared element-wise using an error function~\cite{im2016generating}.
However, the error function assumes the discriminator to be a classifier and therefore this method cannot be applied to Wasserstein-GANs as their discriminator approximates a dual potential.
Instead, we adapted the idea to use the generator of a different model to provide samples for a learned distribution and use the estimated Wasserstein distance as a metric for comparison.
Let $\{(g^{(1)},f^{(1)}),(g^{(2)},f^{(2)}),..., (g^{(n)},f^{(n)})\}$ be a set of Wasserstein-GANs where the $j$-th WGAN's generator is denoted as $g_j$ and its critic is denoted as $f_j$.
Then
\begin{equation}
  W_{i,j} = \mathbb{E}_{x \sim \mathbb{P}_r} [f^{(i)}(x)] - \mathbb{E}_{z \sim p_Z} [f^{(i)}(g^{(j)}((z)))]
\end{equation}
provides an estimate for the Wasserstein distance between $\mathbb{P}_r$ and $\mathbb{P}^{(j)}$ where we use unseen samples from the real data $\mathbb{P}_r$.
The estimate $W_{i,j}$ allows us to draw the following conclusions about the relative generalization capabilities of the Wasserstein-GANs when we compare it to $\hat{W}_{i}$, which is the estimated Wasserstein distance on the training data:
\begin{itemize}
  \item If $W_{i,j} > \hat{W}_i$, the ability of model $i$ to differentiate between the two distributions increases.
  \item If $W_{i,j} < \hat{W}_i$, the ability of model $i$ to differentiate between the distributions decreases.
\end{itemize}
Note that if a Wasserstein-GAN has a Lipschitz constant of $k > 0$, it estimates $k \cdot W_1(\mathbb{P}_r, \mathbb{P}_\theta)$ \cite{arjovsky2017}.
To avoid this scaling error, we define the generalization score for the $i$-th WGAN's discriminator with the $j$-th WGAN's generator as the relative error $W'_{i,j} = (W_{i,j} - \hat{W}_i)/|\hat{W}_i|$.
For a given generator $g_j$ the discriminator $f_{i}$ can better distinguish the data than $f_{i'}$ if $W'_{i,j} > W'_{i',j}$.
An overall generalization score can be computed with $s=\sum_{j=1}^n W'_{i,j}$. 

\subsection{Empirical evaluation}
\begin{figure}
  \centering
  \begin{subfigure}{0.325\linewidth}
    \includegraphics[width=\linewidth]{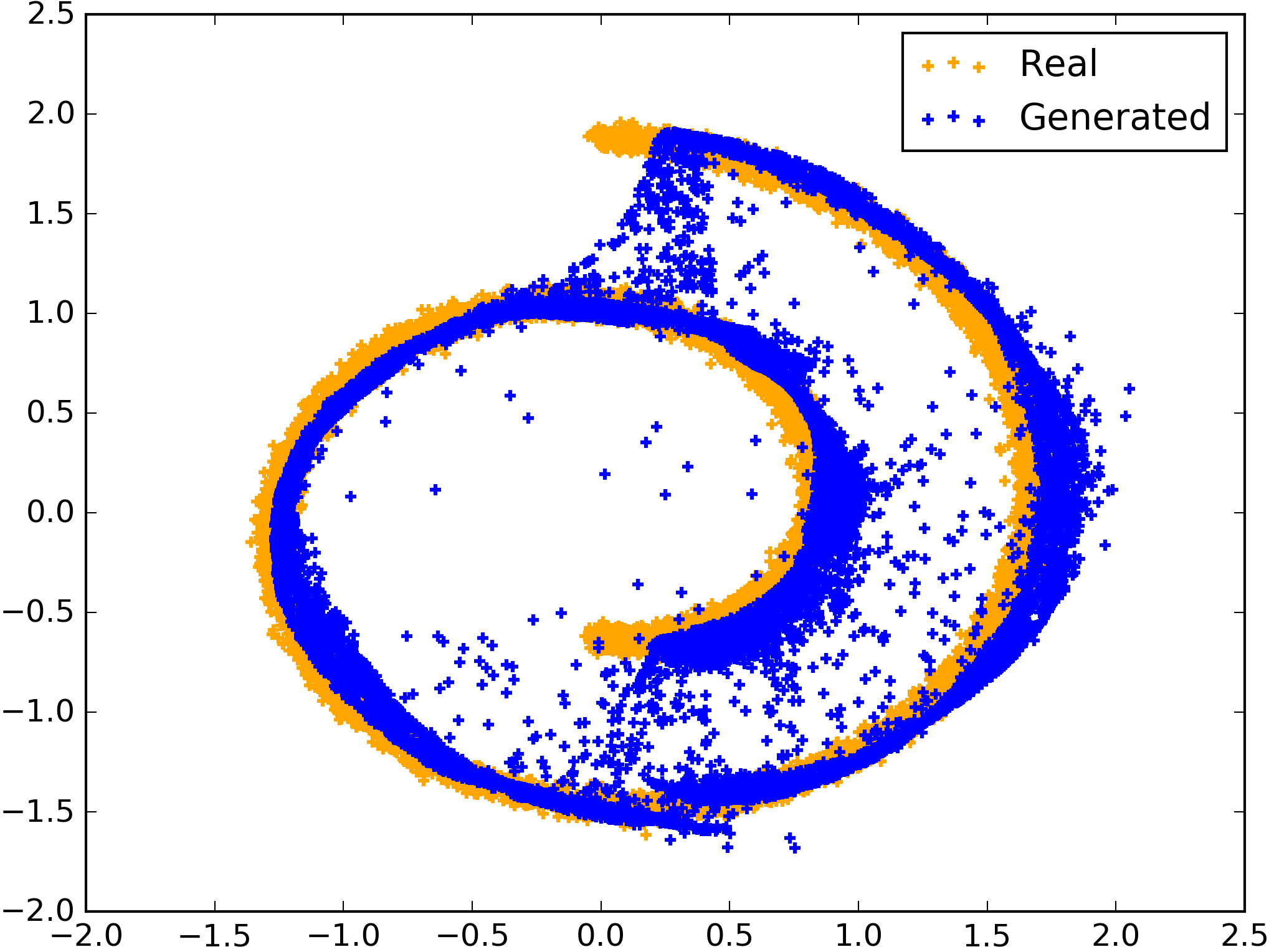} \\
    \caption{WGAN-GP}
  \end{subfigure}
  \begin{subfigure}{0.325\linewidth}
    \includegraphics[width=\linewidth]{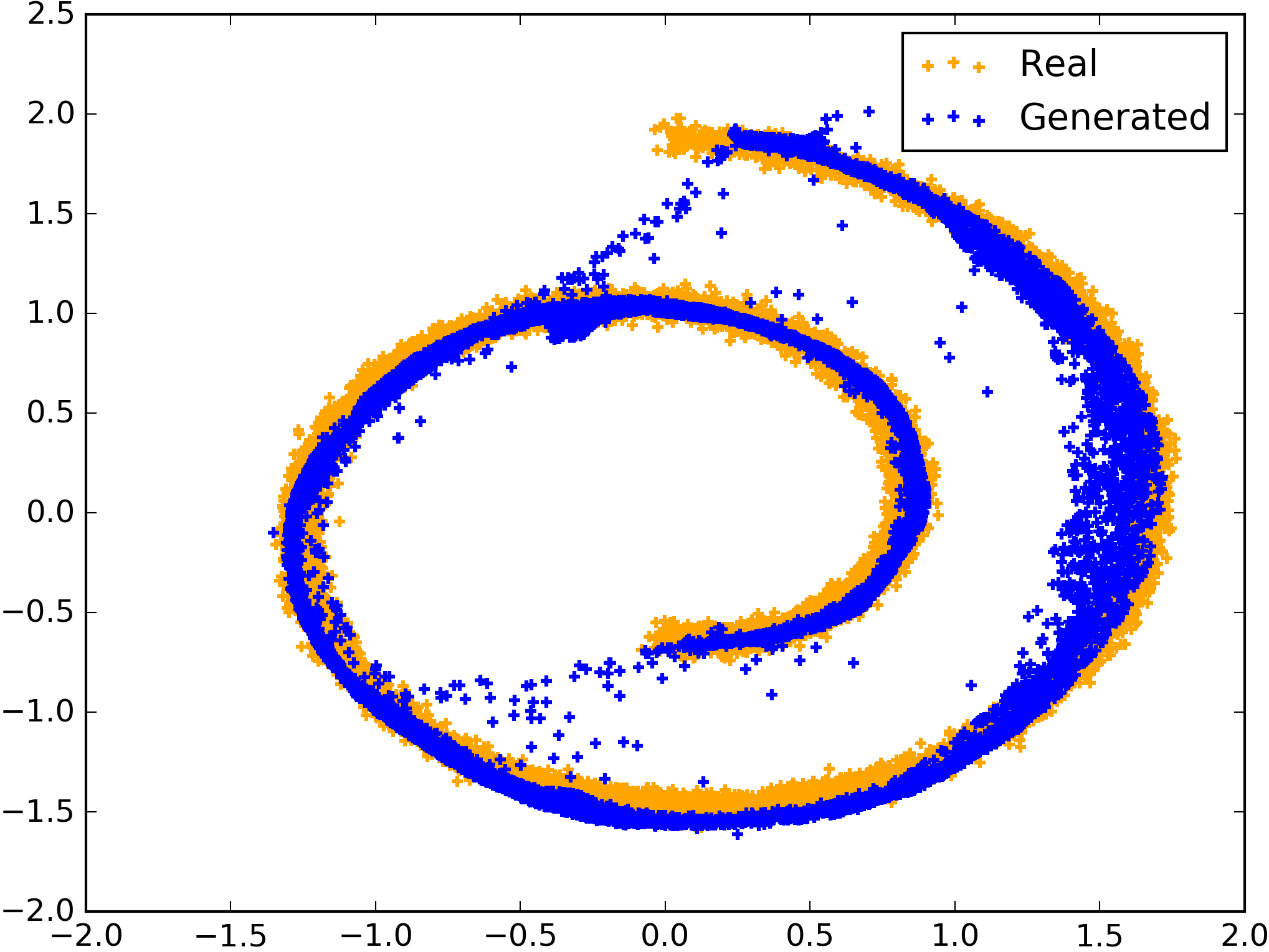} \\
    \caption{WGAN-TTUR}
  \end{subfigure}
  \begin{subfigure}{0.325\linewidth}
    \includegraphics[width=\linewidth]{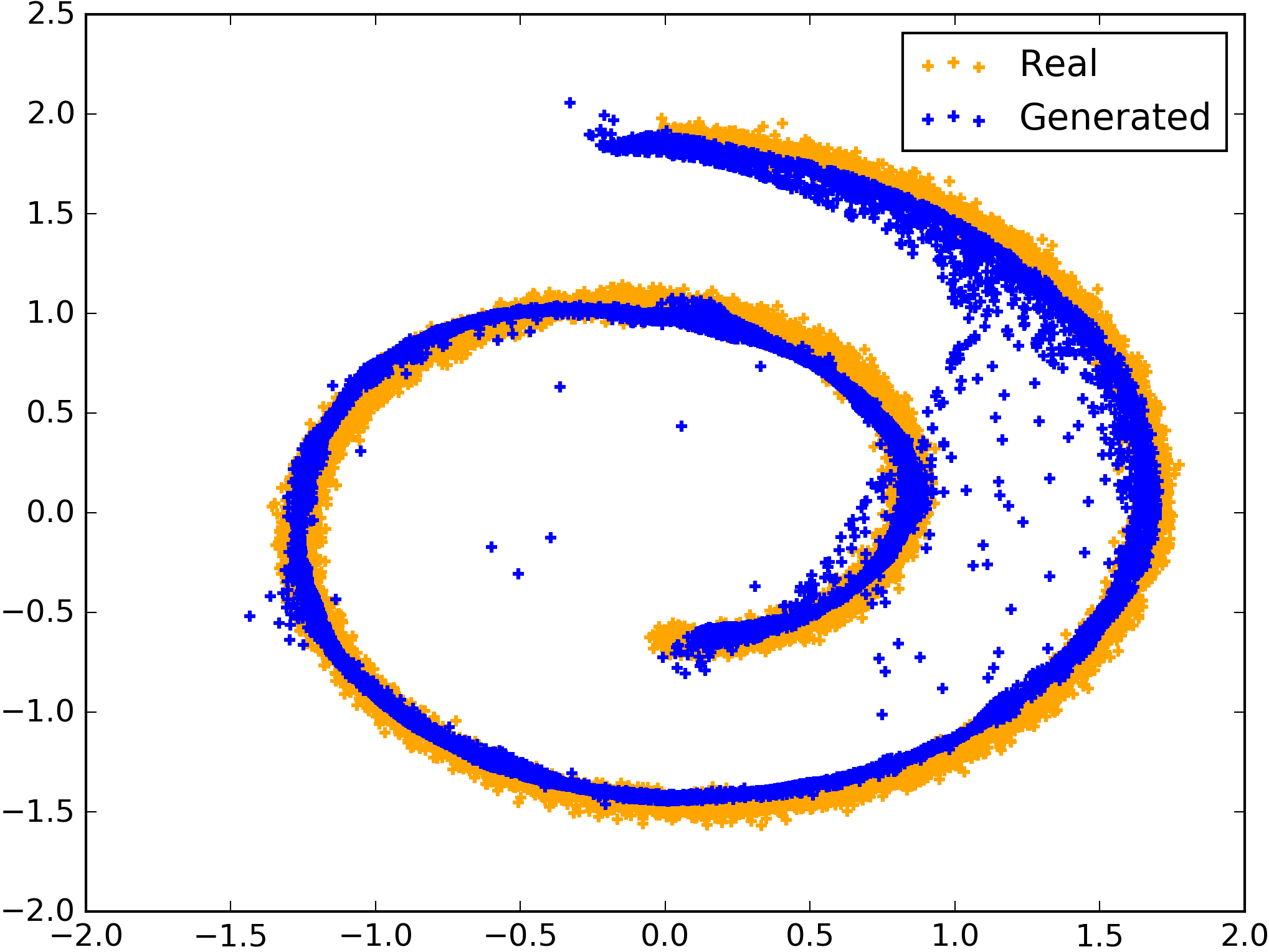} \\
    \caption{Ours}
  \end{subfigure}
  \caption{Learned synthetic data distribution: When trained with the same computational budget, a WGAN trained with our procedure captures the target distribution completely, while the other approaches do not accurately represent the distribution, especially at the ends of the spiral arms.}
  \label{fig:synthetic2}
\end{figure}

To evaluate our approach we compare it to the Wasserstein-GAN with Gradient Penalty (WGAN-GP)~\cite{gulrajani2017improved} to establish a baseline and WGAN-GP trained with a two-time scale update rule as described in \cite{heusel2017gans}, which, to the best of our knowledge, is the state-of-the-art Wasserstein-GAN approach which minimizes the 1-Wasserstein distance without requiring a special architecture.
We consider synthetic distributions as they allow for a more detailed comparison of the captured modes as well as the benchmark dataset CIFAR-10  \cite{krizhevsky2009learning} on which we compute both the models' Inception Score and Fr\'{e}chet Inception Distance.
\begin{figure*}
\centering
\begin{subfigure}{0.21\linewidth}
    \includegraphics[width=\linewidth]{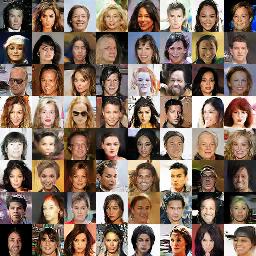} \\
    \caption{}
    \label{fig:biga}
\end{subfigure}
\hspace{0.5in}
\begin{subfigure}{0.28125\linewidth}
    \includegraphics[width=\linewidth]{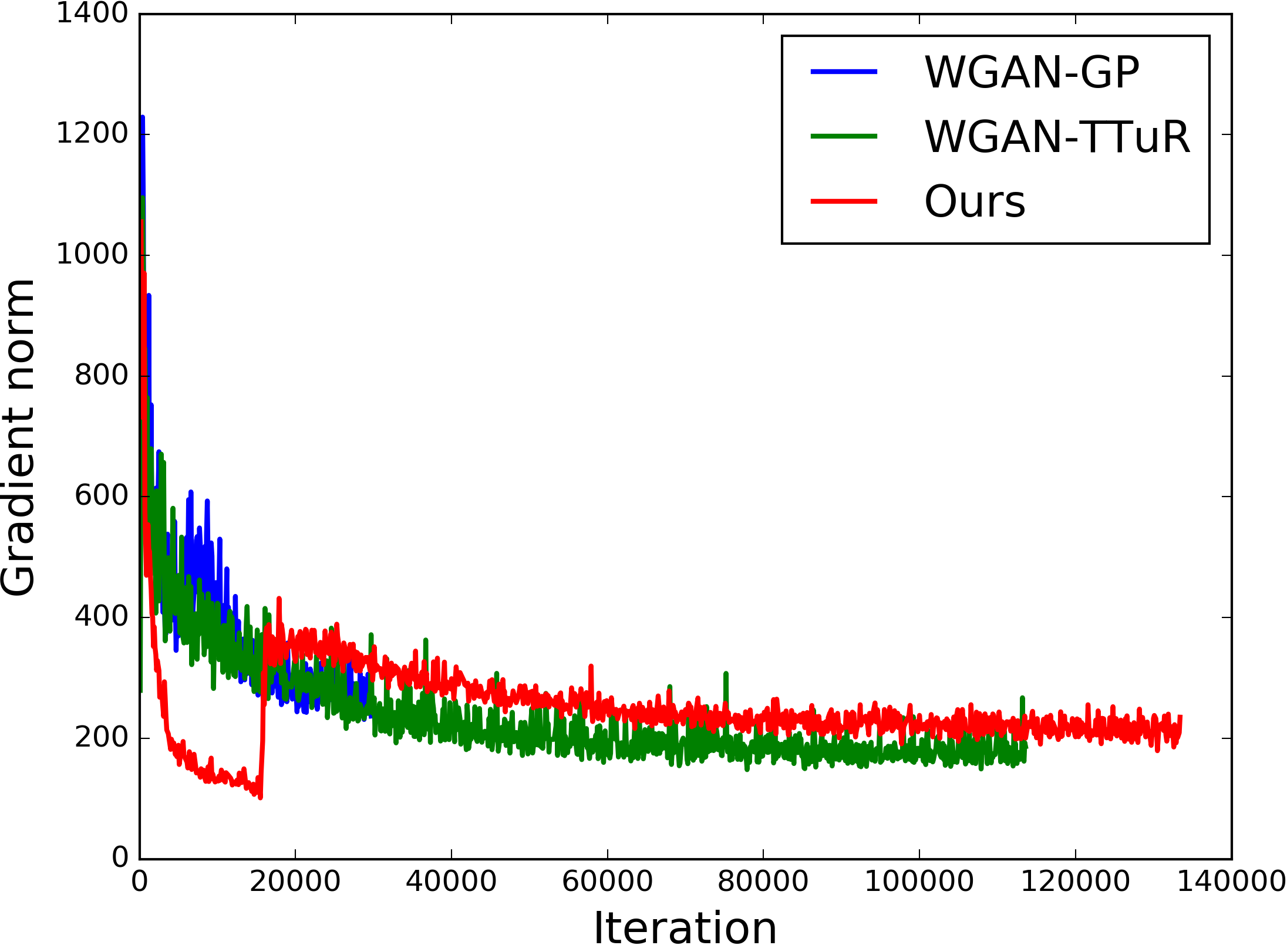} \\
    \caption{}
    \label{fig:bigb}
\end{subfigure}
\hspace{0.5in}
\begin{subfigure}{0.24975\linewidth}
  \includegraphics[width=\linewidth]{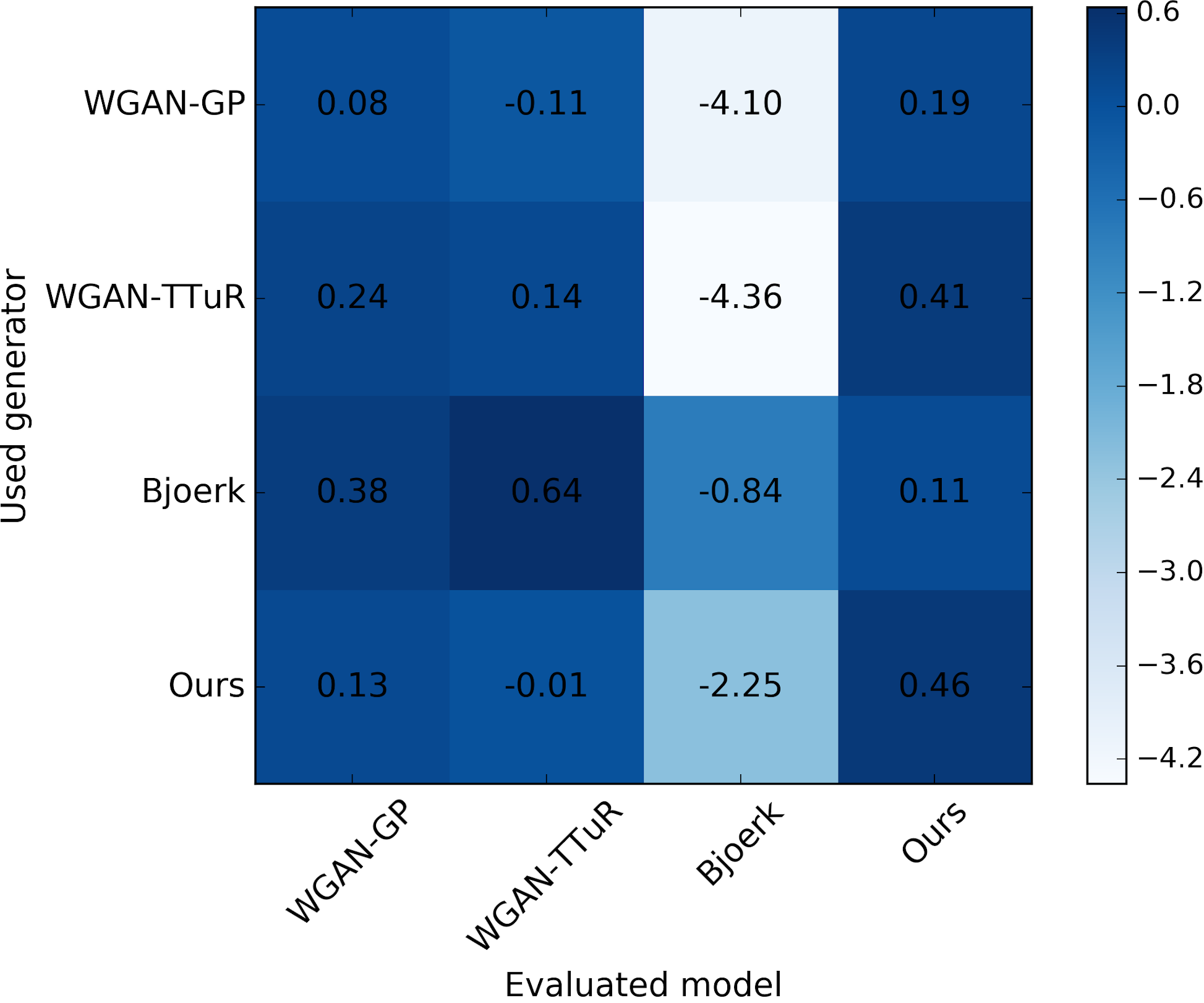} \\
  \caption{}
  \label{fig:bigc}
\end{subfigure}
  \caption{\ref{fig:biga}: Exemplary images generated with our approach. \ref{fig:bigb}: A discriminator of a WGAN trained with our procedure provides a stronger gradient $\nabla_{g(z)} f(g(z))$ to the generator. Note that the procedures are trained with the same computational budget which results in different numbers of iterations. \ref{fig:bigc}: Comparison of the generalization capabilities of Wasserstein-GAN discriminators based on our new metric. A higher value is better.}
\label{fig:big}
\end{figure*}
\textbf{Datasets, architecture and parameters:}
To learn the synthetic distribution, we use a $4$-layer MLP with linear outputs to represent both the generator and the discriminator.
Furthermore, we use Rectified Linear Units (ReLU) as activations for the hidden layers and do not consider additional normalizations or constraints in the network.
For image datasets, we use a convolution architecture based on the DCGAN \cite{radford2015unsupervised}.
For WGAN-GP and WGAN-TTUR, we replaced the batch normalization in the discriminator with layer normalization \cite{ba2016layer} as recommended in \cite{salimans2016improved}. 
On the synthetic dataset we trained all models for 10 minutes with a batchsize of $128$ and on CIFAR-10 all models were trained for 60 minutes with a batchsize of $64$ on a Nvidia GTX 1080.
For WGAN-GP and WGAN-TTUR, we used the hyper-parameters provided in the original publications.
\textbf{Comparison}
The visualization of samples drawn from a synthetic distribution and samples generated by Wasserstein-GANs trained with different procedures are visualized in Figure \ref{fig:synthetic2}.
Both WGAN-GP and WGAN-TTUR do not accurately represent the ends of the spiral arms, while samples generated by our method completely cover the target distribution.

In Table \ref{tab:scores2}, we report the  Fr\'{e}chet Inception Distance and Inception Score of the different procedures for the CIFAR-10 dataset.
In addition, we also show the number of iterations per second during the training process for each of the methods.
Our method outperforms the other methods with respect to the Fr\'{e}chet Inception Distance, while also outperforming WGAN-GP with respect to the model's Inception Score.
Note that our method additionally offers the highest computational efficiency.
\begin{table}
  \caption{Inception Score (IS) and Fr\'{e}chet Inception Distance (FID) on CIFAR-10. A higher IS is better; a lower FID is better.}
  \label{tab:scores2}
	\scriptsize
  \begin{center}
    \begin{tabular}{|c|c|c|c|} \hline
       Model         & FID                      & IS                      & $\frac{\textrm{Iterations}}{\textrm{sec}}$ \\ \hline
       Wgan-GP       & $40.35 \pm 0.1$          & $6.10 \pm 0.06$         & $8.30$ \\ \hline
       Wgan-TTuR     & $37.11 \pm 0.05$         & $\mathbf{6.8 \pm 0.05}$ & $31.63$ \\ \hline
       Standard reg. & $67.75 \pm 0.17$         & $4.80 \pm 0.06$         & $\mathbf{51.49}$  \\ \hline
			 Cayley reg.   & $65.175 \pm 0.10$        & $4.69 \pm 0.02$         & $14.70$ \\ \hline
       Iterative reg.& $43.18 \pm 0.05$         & $5.85 \pm 0.04$         & $51.23$  \\ \hline
       \textbf{Ours} & $\mathbf{35.22 \pm 0.1}$ & $6.50 \pm 0.04$         & $37.15$ \\ \hline
    \end{tabular}
  \end{center}
\end{table}
Comparing the estimated Wasserstein-distance using our new metric in Figure \ref{fig:bigc}, we observe that our proposed method has the highest overall generalization score of $s=1.17$ while the next best model WGAN-GP only reaches $s=0.83$.
As the diagonal reflects the discriminators' overfitting with respect to the test data it is of special interest and our method achieves the highest performance in distinguishing generated data from unseen real data with a score of $W'_{4,4}=0.46$.

\begin{figure}
  \centering
  \begin{subfigure}{0.325\linewidth}
    \includegraphics[width=\linewidth]{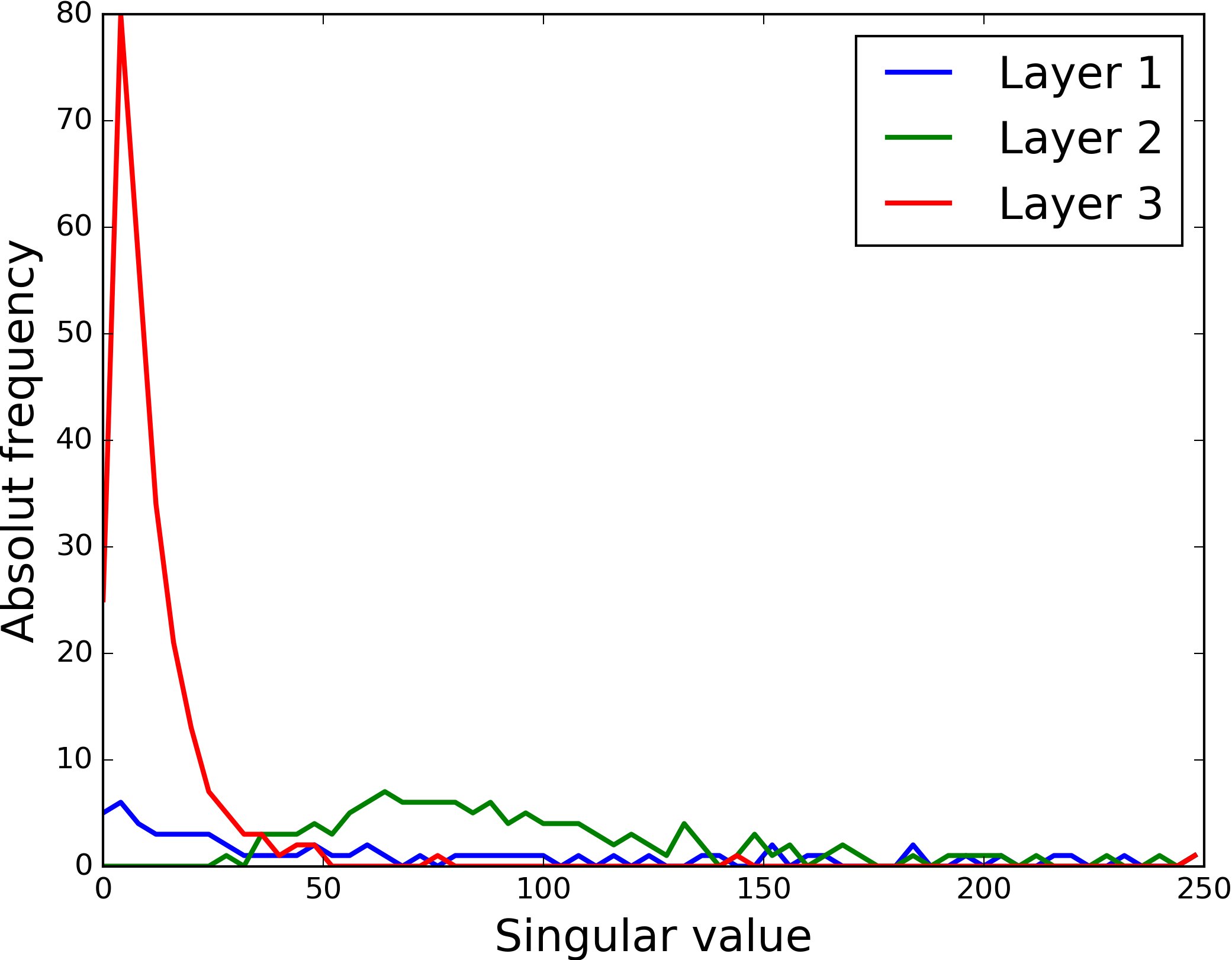} \\
    \caption{WGAN-TTuR}
  \end{subfigure}
  \hspace{0.5in}
  \begin{subfigure}{0.325\linewidth}
    \includegraphics[width=\linewidth]{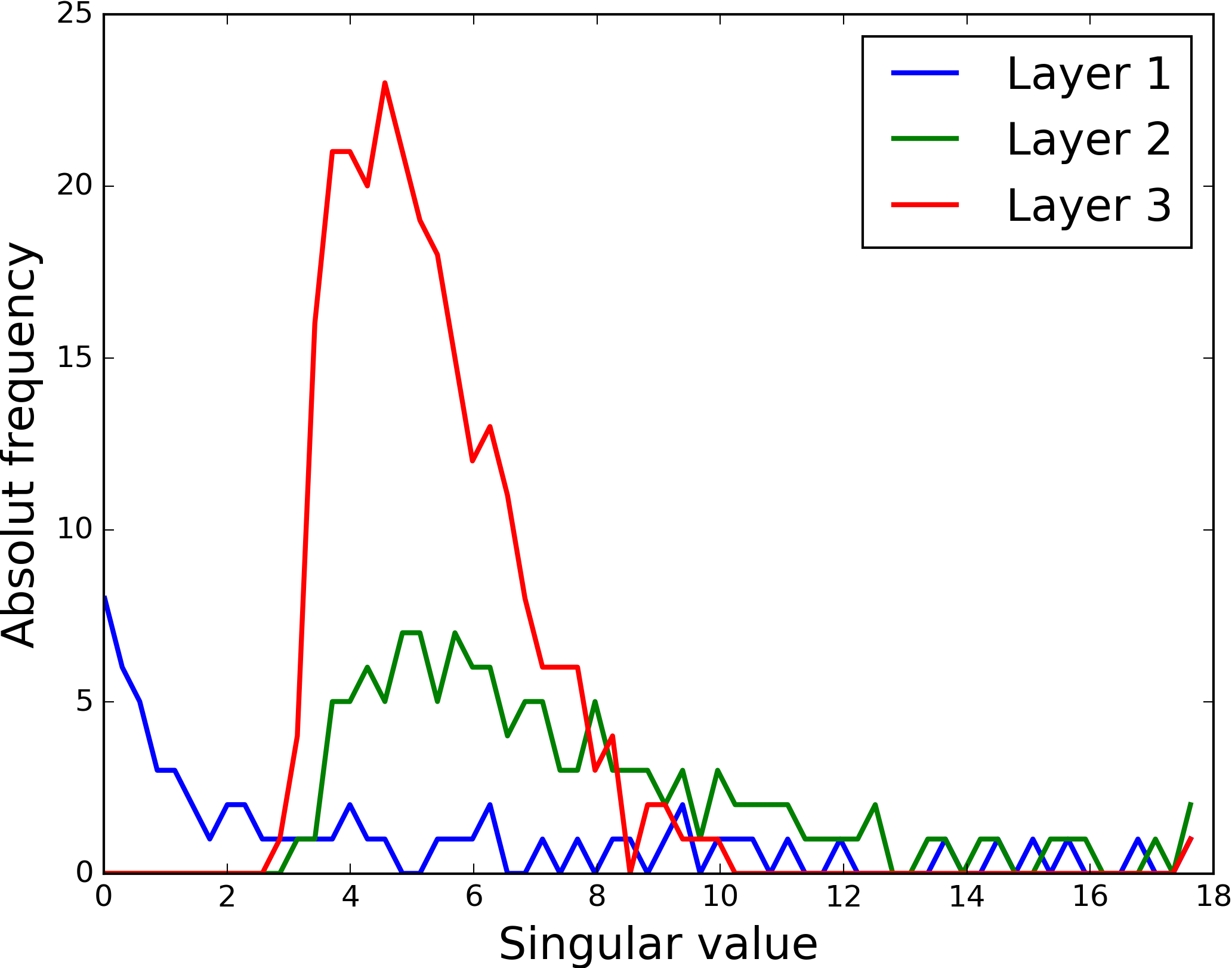} \\
    \caption{Ours}
  \end{subfigure}
  \caption{Distribution of specular-values in the discriminator's convolution layers after being trained on CIFAR-10. Note the different scale is a result of the uneven distribution in WGAN-TTUR.}
  \label{fig:singluar}
\end{figure}

To further compare the different procedures for the training of Wasserstein-GANs and to gain additional insights regarding the benefits of our method, we plot the discriminators' gradient norm in Figure \ref{fig:bigb}.
The sudden increase in the gradient norm is a result from relaxing the orthogonality constraint. 
As that the generator learns to minimize $E[f_\omega(g_\theta(z))]$, the gradient norm of $\nabla_{g_\theta(z)} E[f_\omega(g_\theta(z))]$ is crucial during training.
In general, our procedure provides a stronger gradient to the generator for the majority of iterations when compared to WGAN-TTUR.
Furthermore the gradient is more stable than the one of the competing techniques as it shows the lowest amount of noise over the iterations, even though all models have been trained with the same batchsizes.
An additional benefit of our method is a more even distribution of the weights' spectral-values in the discriminator as shown in Figure \ref{fig:singluar}.
As argued in \cite{miyato2018spectral}, a more even distribution of spectral values encourages the discriminator to capture more features of the real dataset.

\section{Conclusion}
In this work, we outlined a connection between the orthogonal weight matrices in neural networks and the Lipschitz continuity required by Wasserstein-GANs.
We have empirically investigated the possibility of replacing the gradient norm regularization by different orthogonalization methods.
We found the training with hard constraint orthogonalization methods to be stable and that all considered orthogonalization methods are able to enforce the Lipschitz constraint.
However, the learned distributions did not exhibit the same fidelity as the distributions learned by established training methods.
Based on the insights gained from this investigation, we proposed a new trainings method which utilizes the increased stability but avoids restricting the model's capacity.
Finally, we were able to demonstrate that a Wasserstein-GAN discriminator trained with this procedure has an increased generalization capability and its weight matrices exhibit more evenly distributed singular-values, which enables the model to better represent the target distribution.

{\small
\bibliographystyle{ieee_fullname}
\bibliography{report}
}
\clearpage
\appendix
\part*{Appendix}
\FloatBarrier
\section{Demonstration of problems in conjunction with gradient norm penalties}

\begin{figure}
  \centering
  \begin{subfigure}{0.49\linewidth}
    \includegraphics[width=\linewidth]{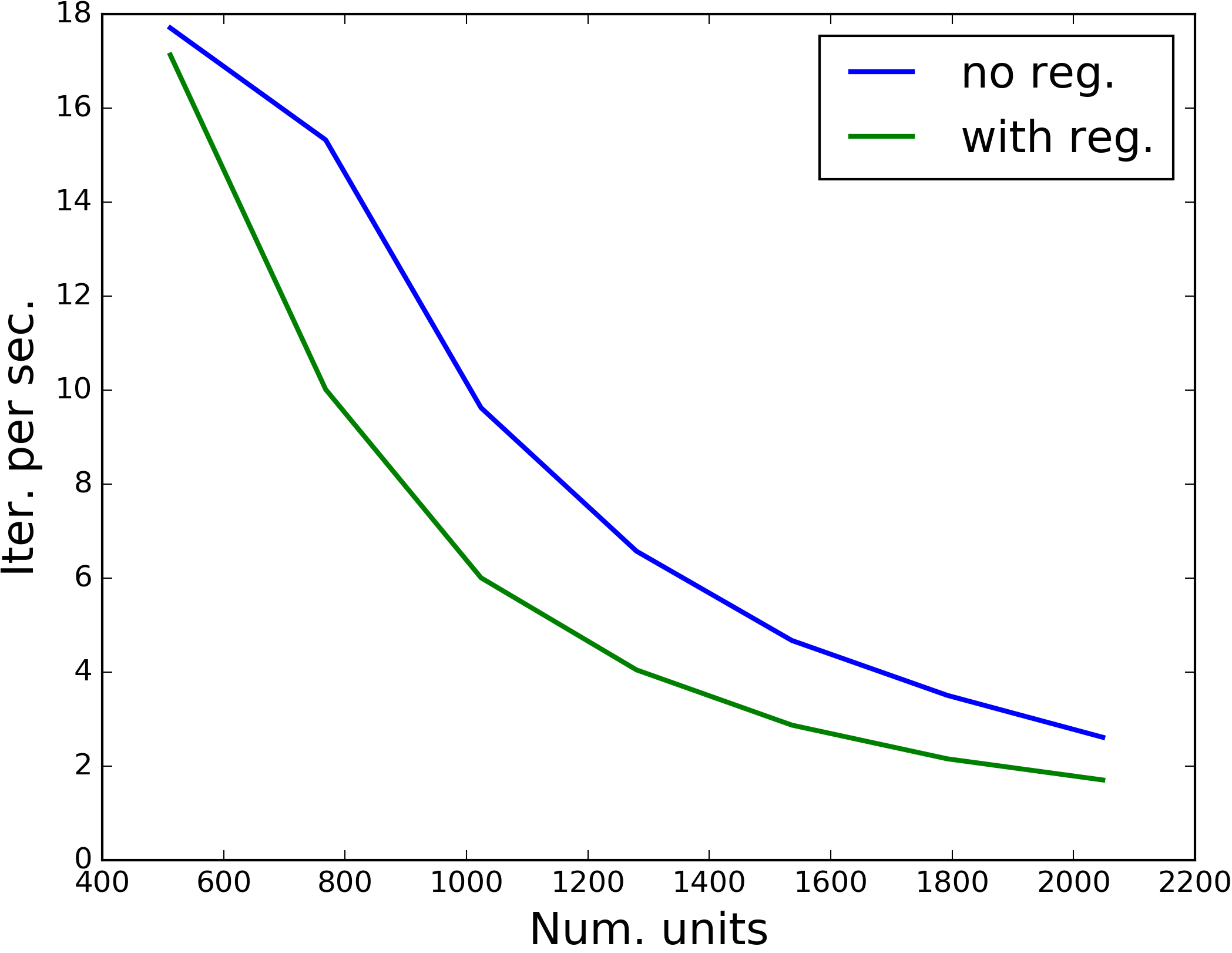} \\
    \caption{Scaling with number units}
  \end{subfigure}
  \begin{subfigure}{0.49\linewidth}
    \includegraphics[width=\linewidth]{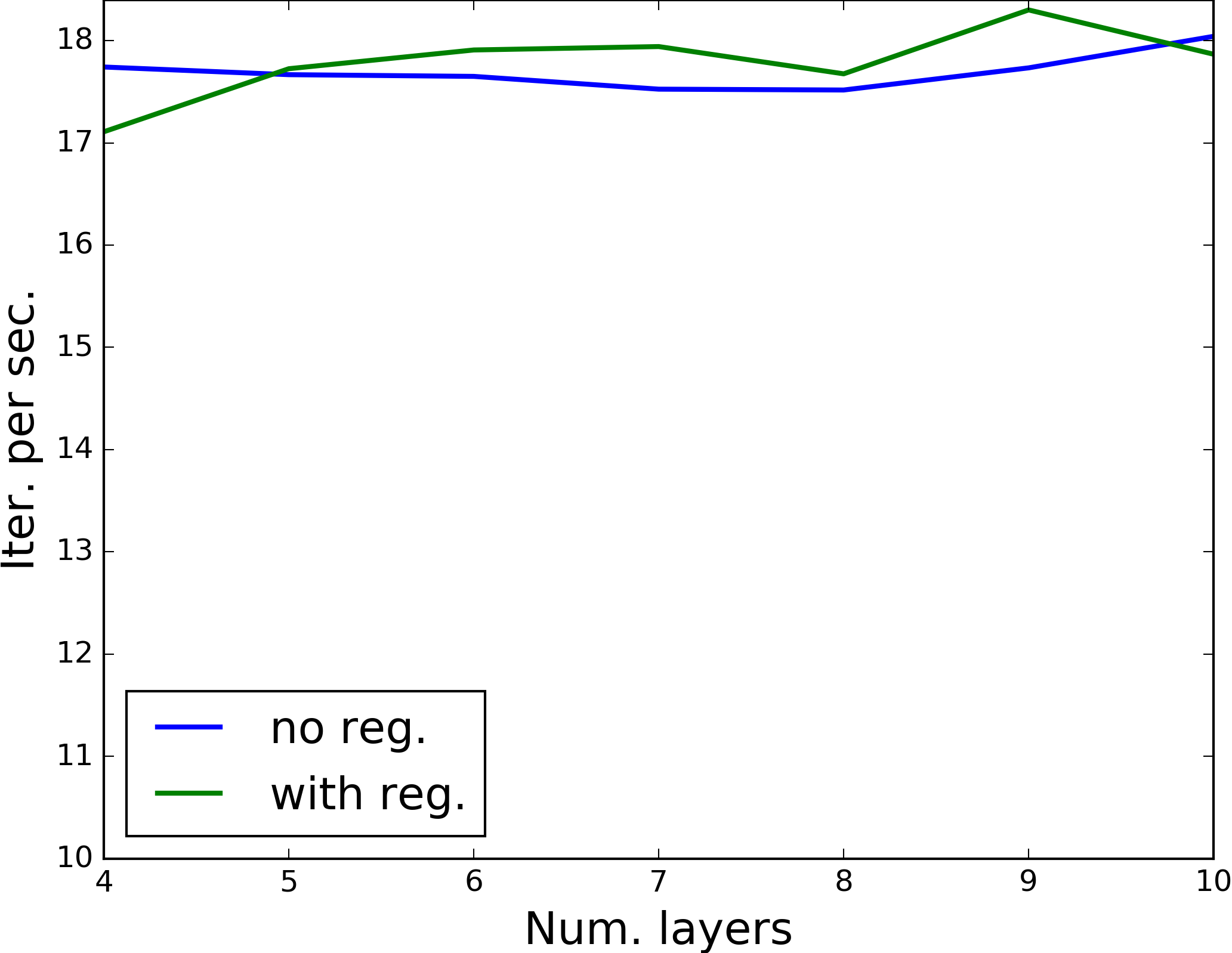} \\
    \caption{Scaling with layers}
  \end{subfigure}
  \caption{Number of iterations per second during training for a Wasserstein-GAN trained with and without gradient normalization according to \cite{gulrajani2017improved}.}
  \label{fig:runtime}
\end{figure}

\begin{figure}
  \centering
  \begin{subfigure}{0.325\linewidth}
    \includegraphics[width=\linewidth]{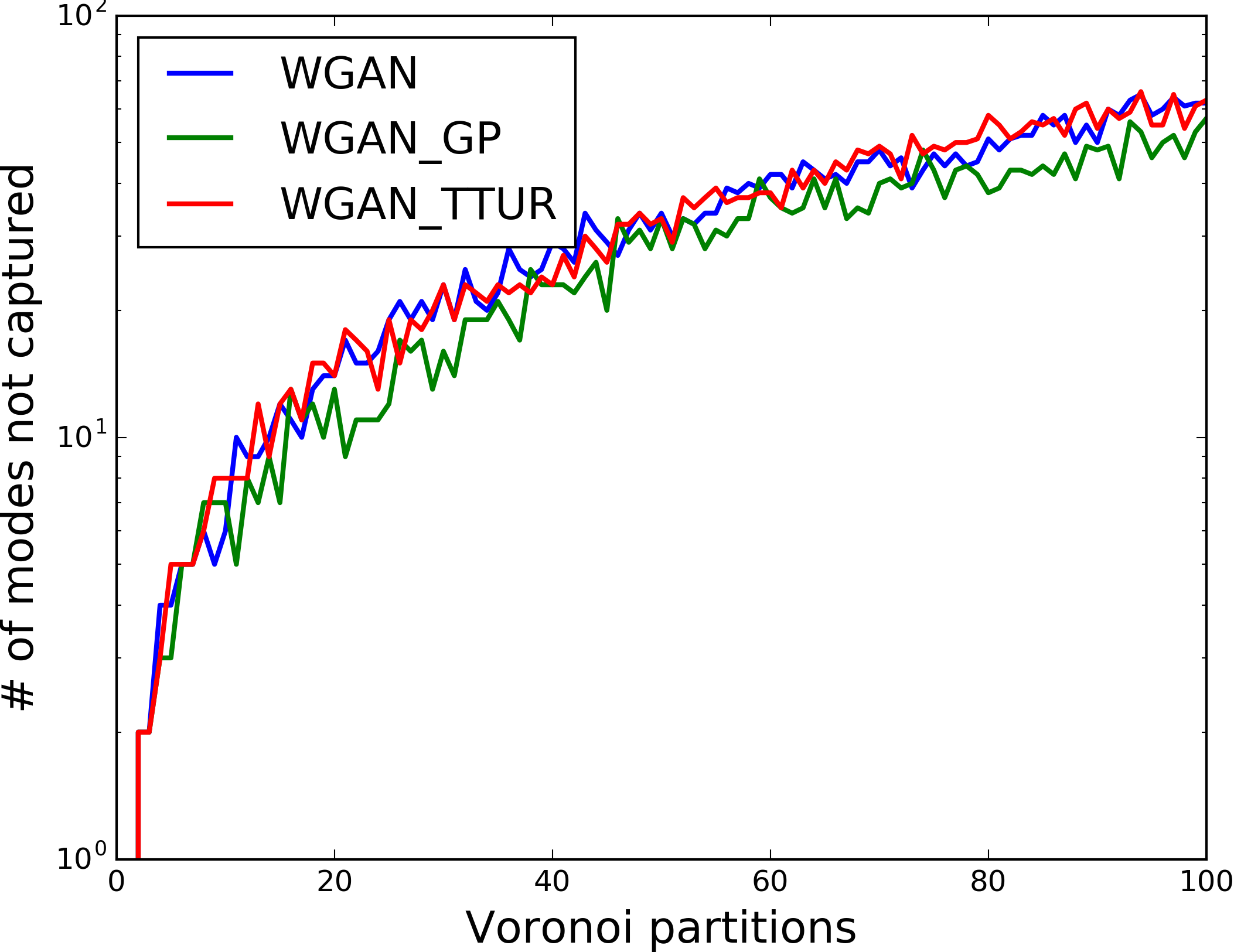} \\
    \caption{MNIST}
  \end{subfigure}
  \begin{subfigure}{0.325\linewidth}
    \includegraphics[width=\linewidth]{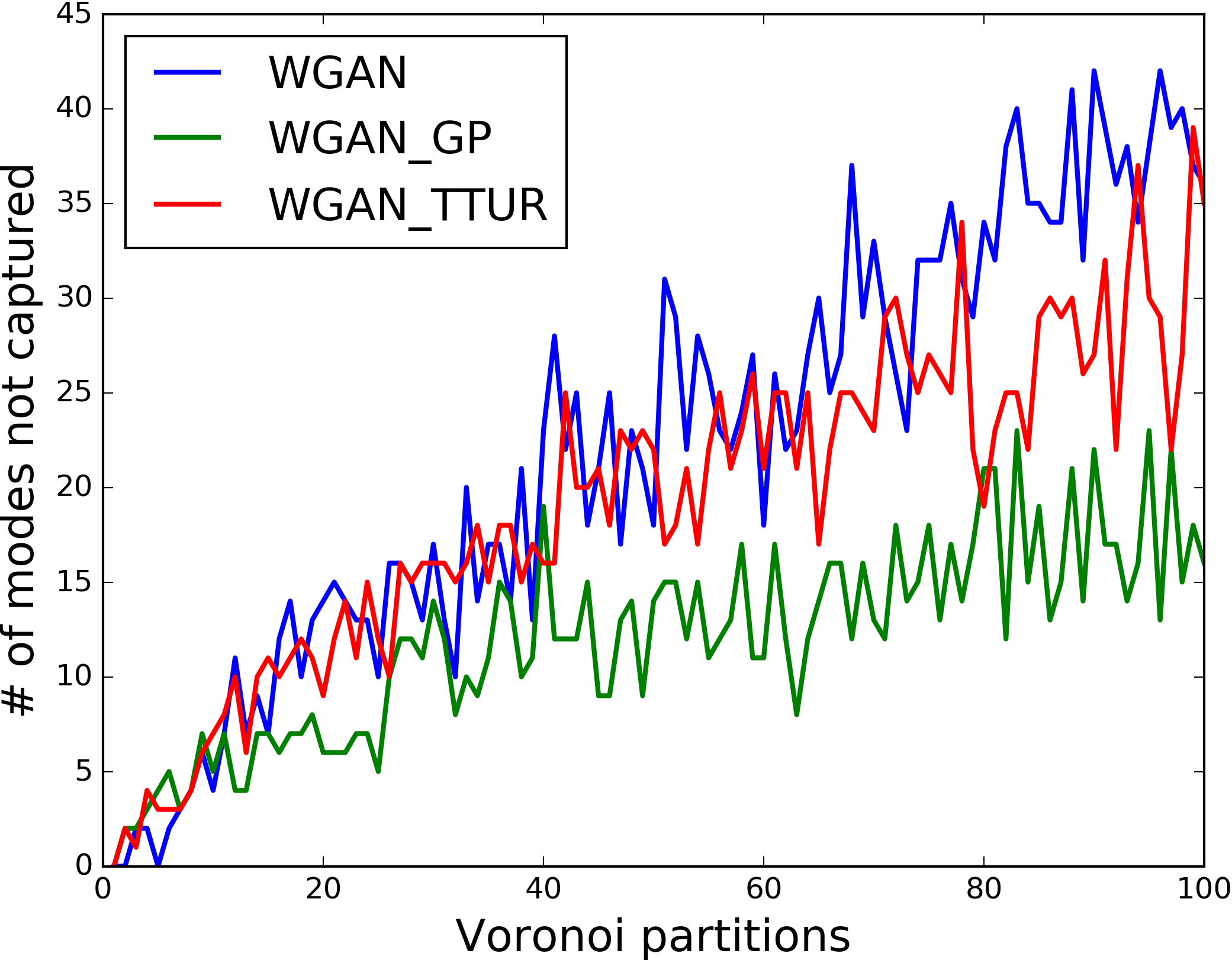} \\
    \caption{CIFAR-10}
  \end{subfigure}
  \begin{subfigure}{0.325\linewidth}
    \includegraphics[width=\linewidth]{/motivation/modes_celeba.png} \\
    \caption{CelebA}
  \end{subfigure}
  \caption{Number of approximated modes which are statistically significantly less preserved in the generated distribution when compared to its test dataset.}
  \label{fig:modes}
\end{figure}

\paragraph{Influence of gradient norm regularization on the runtime}
We demonstrate the increase in computational complexity by training a Wasserstein-GAN with weight clipping and a Wasserstein-GAN with gradient normalization on a synthetic dataset.
The used architecture is a multilayer perceptron (MLP) where we vary either the number of layers while keeping the number of (hidden) units fixed or vice versa.
We vary the number of layers in the range of $n \in \{4,..., 10\}$ and the number of units per hidden layer $m \in \{512, 768, ..., 2048\}$.
Default values when varying the other parameter are $n = 4$ layers and $m = 512$ units.
The results in Figure \ref{fig:runtime} show that the number of iterations per second decreases by up to 30\% with gradient regularization.
While an increase in the number of layers with a small but constant overall number of units in the MLP does not change the computational efficiency,
we observe a decrease in the number of iterations per second during training when the number of units is increased.
This decrease is significantly larger when using regularization and combined with multiple discriminator updates per generator update  solves down the training of WGAN-GP.
\paragraph{Analysis of mode preservation for different WGAN approaches}
One of the main benefits of Wasserstein GANs over standard GANs is their capability to mitigate mode collapse.
However, applying techniques such as the TTUR~\cite{heusel2017gans} for reducing the training time weaken this effect.
To compare the mode collapse for different established Wasserstein-GAN approaches, we trained a Wasserstein-GAN with weight clipping (WGAN), a Wasserstein-GAN with gradient penalty (WGAN-GP) and a Wasserstein-GAN with TTUR (WGAN-TTUR) using the architecture described in the accompanying paper on the MNIST~\cite{lecun1998mnist}, CIFAR-10~\cite{krizhevsky2009learning} and CelebA~\cite{Liu:2015:DLF} datasets.
Each of the models was trained for $10^5$ iterations using the the hyper-parameters provided in the original publications. 
Finally, we evaluated the mode-collapse using the procedure proposed by Richardson and Weiss~\cite{richardson2018gans}.
The results in Figure \ref{fig:modes} demonstrate that WGAN-TTUR has a significantly lower number of represented modes when compared to WGAN-GP. 
In turn, WGAN-GP outperforms the standard WGAN.

\section{Additional implementation details of the proposed method}

\paragraph{Initialization}
The initialization of network weights has been studied extensively and it has been demonstrated that a careful initialization already improves a network's performance significantly \cite{Xie:2017}.
Inspired by the initialization proposed by Saxe et al.~\cite{Saxe14exactsolutions}, we initialize the weights by computing an SVD $M = U \Sigma V^T$, replacing all singular values $\sigma_i$ by $\lambda$ and setting the weights to $W = U \text{diag}(\lambda, ..., \lambda) V^T$.
We found it to be beneficial to further relax the orthogonality constraint by setting $\lambda > 1$.
To motivate this parameter choice, we consider the derivative of the generator's objective function $\mathbb{E}_{z \sim \mathcal{Z}} \left[ f_\omega(g_\theta(z)) \right] $.
Note that its gradient can be written as $\mathbb{E}_{z \sim \mathcal{Z}} \left[ \nabla_\theta f_\omega(g_\theta(z)) \right]$ \cite{arjovsky2017} and that the chain rule implies that this gradient can be factorized into a product between $\nabla_x f(x)$ with $x = g_\theta(z)$ and $\nabla_\theta g_\theta(z)$.
If we recall equation 3 from the accompanying paper, we see that there is a direct connection between the gradient norm of $f$ and the gradient norm of the generator's objective, and, as the $2$-norm of a matrix is its largest singular value, we can increase the generator's training speed by scaling the singular values.

\begin{figure}
  \begin{center}
    \includegraphics[width=0.75\linewidth]{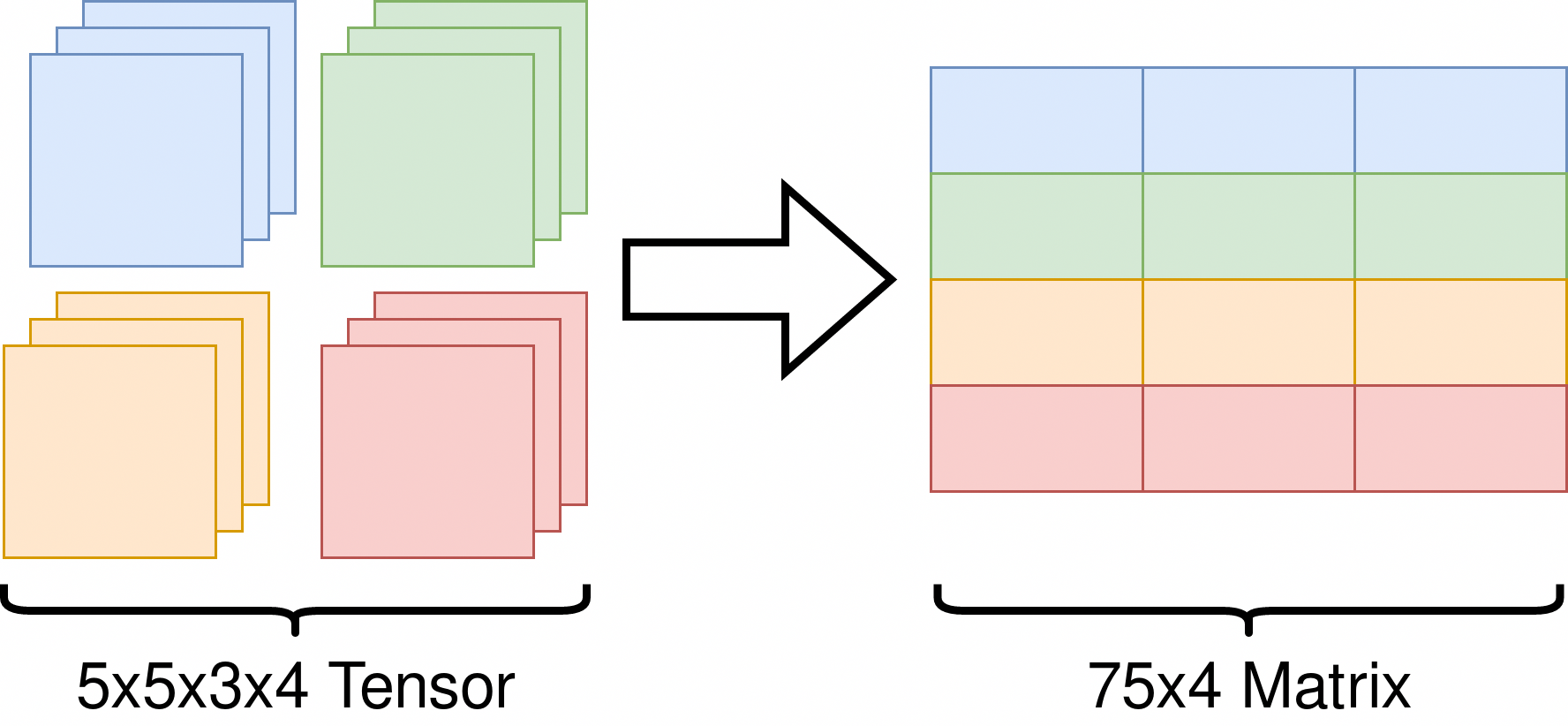}
    \caption{Reshaping the mode $n$ tensor fibers into a matrix.}
    \label{fig:reshaping}
  \end{center}
\end{figure}

\paragraph{Extension to convolutions}
We assumed that the discriminator is a feed-forward network build from linear operations $L = W x$ and $1$-Lipschitz continuous activations.
However, GANs are predominantly used in image-based applications which heavily rely on network architectures that are based on convolution operations.
Convolution operations can be unrolled into a linear operation.
While this procedure is correct from a theoretical perspective, the resulting matrix would be too large to train a complex network in reasonable time.
In order to avoid this problem, we extend the procedure by constructing a matrix from the modes of a tensor.
Let $W \in \mathbb{R}^{n \times m \times l \times k}$ be the $4D$-tensor representing a discrete convolution with a filter size of $n \times m$, where $l$ denotes the number of filters of the previous layer and the $k$ the number of output filters.
Instead of unrolling the operation, we reshape the tensor into a matrix by flattening each kernel into an $n \times m \times l$ row vector and concatenating the resulting row vectors vertically into a matrix with dimensions $(n \cdot m \cdot l) \times k$.
An exemplary illustration of this tensor reshaping is shown in Figure \ref{fig:reshaping}.

\begin{figure}
  \centering
  \begin{subfigure}{0.325\linewidth}
    \includegraphics[width=\linewidth]{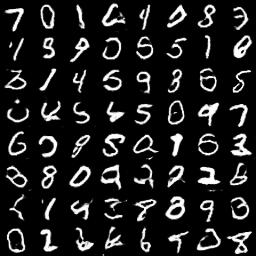} \\
    \caption{WGAN-GP}
  \end{subfigure}
  \begin{subfigure}{0.325\linewidth}
    \includegraphics[width=\linewidth]{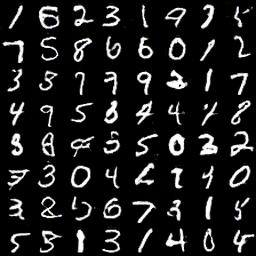} \\
    \caption{WGAN-TTUR}
  \end{subfigure}
  \begin{subfigure}{0.325\linewidth}
    \includegraphics[width=\linewidth]{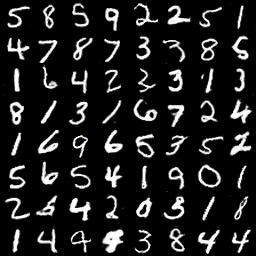} \\
    \caption{Ours}
  \end{subfigure}
  \caption{Samples generated by the model trained on the MNIST dataset resized to be $32 \times 32$ pixels. The samples were chosen at random.}
  \label{fig:mnist}
\end{figure}
\begin{figure}
  \centering
  \begin{subfigure}{0.325\linewidth}
    \includegraphics[width=\linewidth]{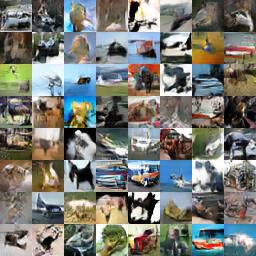} \\
    \caption{WGAN-GP}
  \end{subfigure}
  \begin{subfigure}{0.325\linewidth}
    \includegraphics[width=\linewidth]{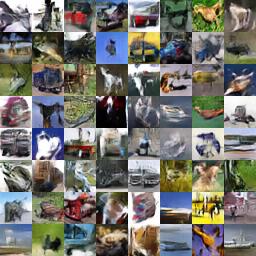} \\
    \caption{WGAN-TTUR}
  \end{subfigure}
  \begin{subfigure}{0.325\linewidth}
    \includegraphics[width=\linewidth]{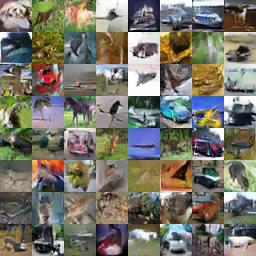} \\
    \caption{Ours}
  \end{subfigure}
  \caption{Samples generated by the model trained on the CIFAR-10 dataset. The samples were chosen at random.}
  \label{fig:cifar10}
\end{figure}
\begin{figure}
  \centering
  \begin{subfigure}{0.325\linewidth}
    \includegraphics[width=\linewidth]{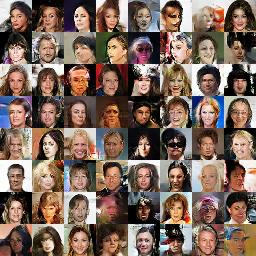} \\
    \caption{WGAN-GP}
  \end{subfigure}
  \begin{subfigure}{0.325\linewidth}
    \includegraphics[width=\linewidth]{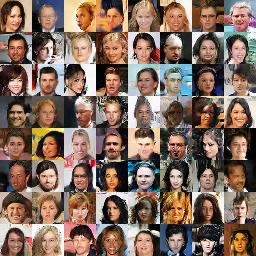} \\
    \caption{WGAN-TTUR}
  \end{subfigure}
  \begin{subfigure}{0.325\linewidth}
    \includegraphics[width=\linewidth]{/evaluation/wgan_init_images.jpg} \\
    \caption{Ours}
  \end{subfigure}
  \caption{Samples generated by the model trained on the CelebA dataset resized to be $32 \times 32$ pixels. The samples were chosen at random.}
  \label{fig:celeba}
\end{figure}
\begin{figure}
  \centering
  \begin{subfigure}{0.49\linewidth}
    \includegraphics[width=\linewidth]{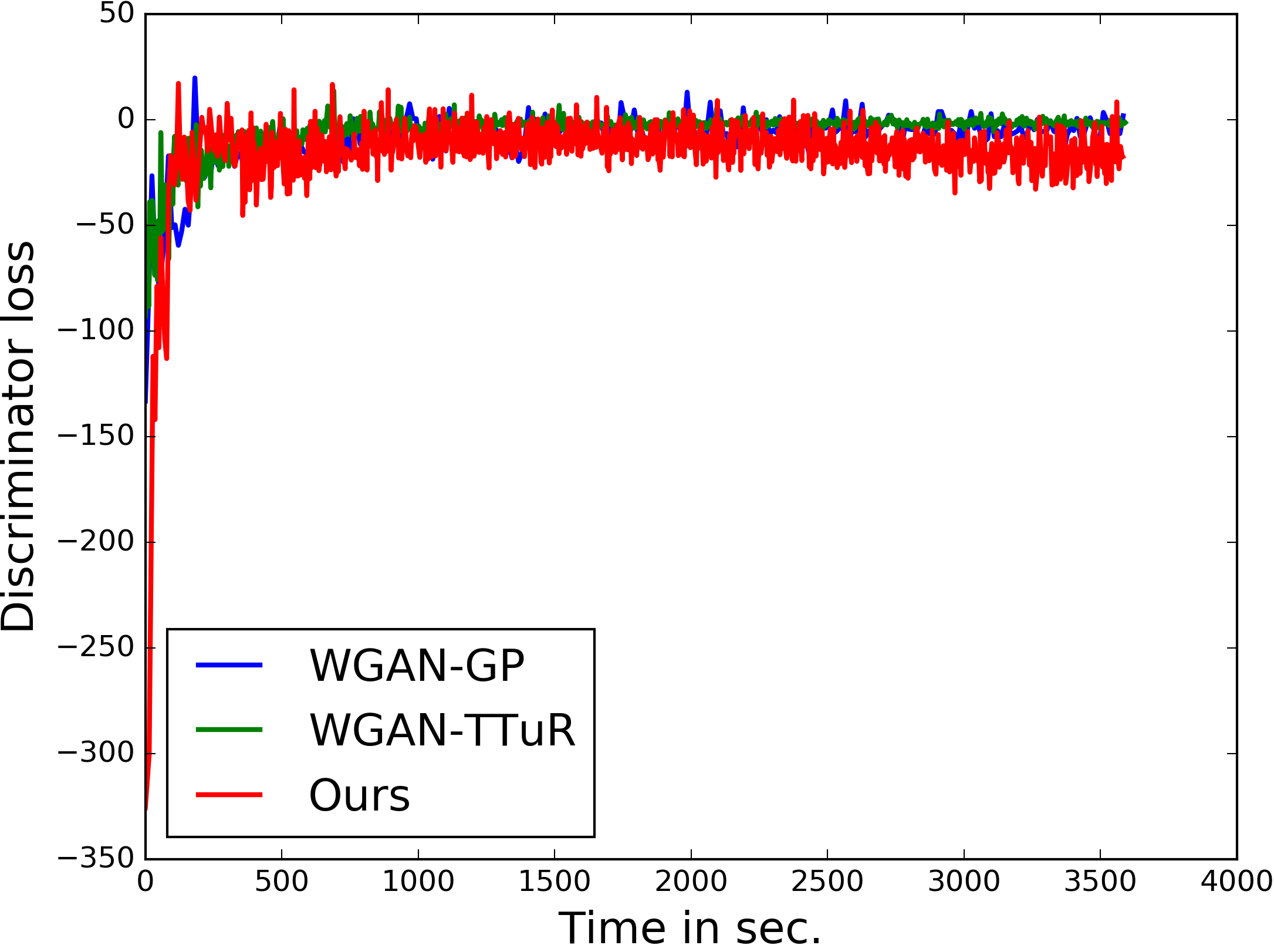} \\
    \caption{Discriminator loss}
  \end{subfigure}
  \begin{subfigure}{0.49\linewidth}
    \includegraphics[width=\linewidth]{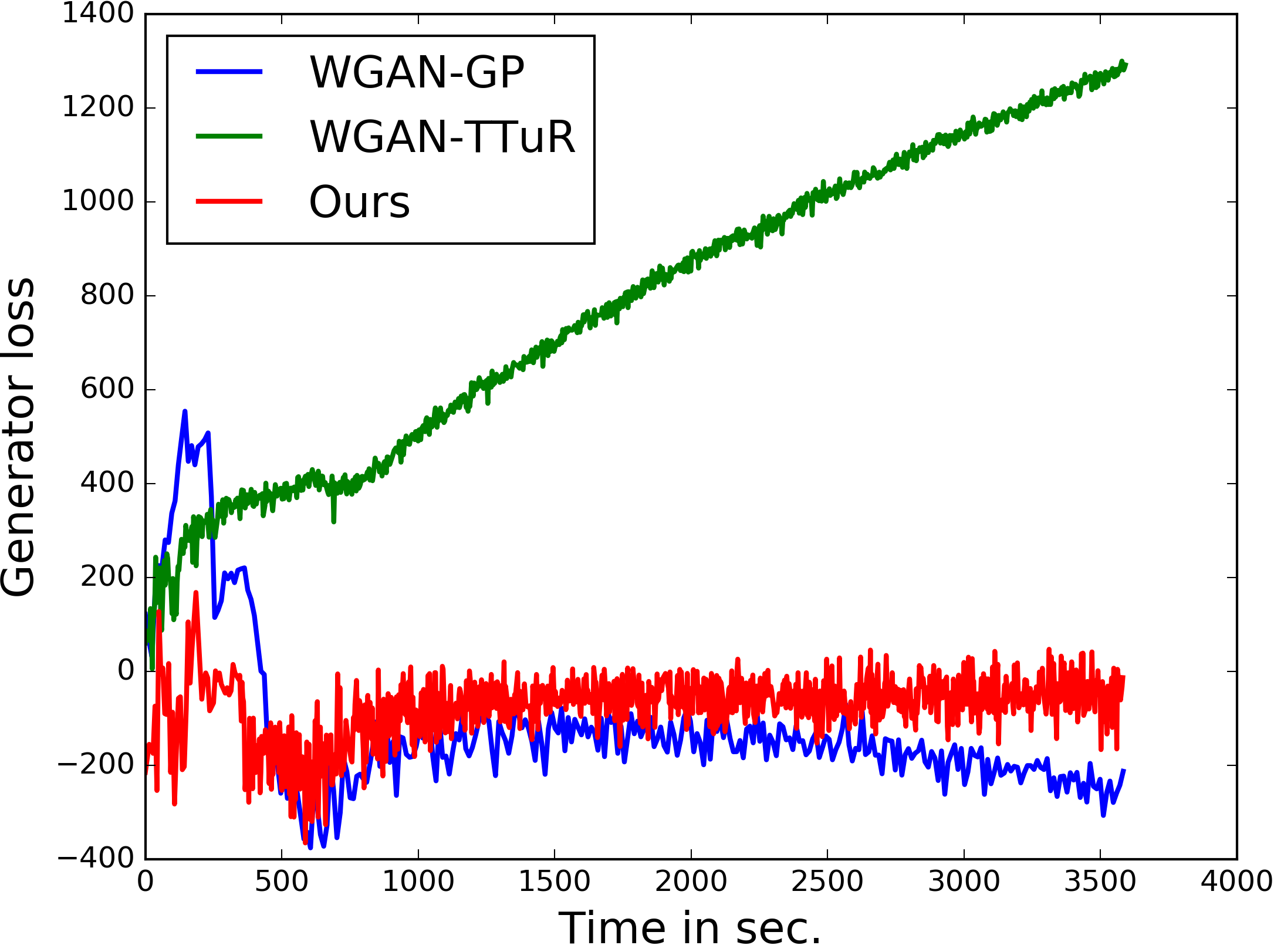} \\
    \caption{Generator loss}
  \end{subfigure}
  \caption{Wall-clock aligned discriminator and generator loss curves, which resulted from training the models on CIFAR-10.}
  \label{fig:cifar10_losses}
\end{figure}
\begin{figure}
  \centering
  \begin{subfigure}{0.49\linewidth}
    \includegraphics[width=\linewidth]{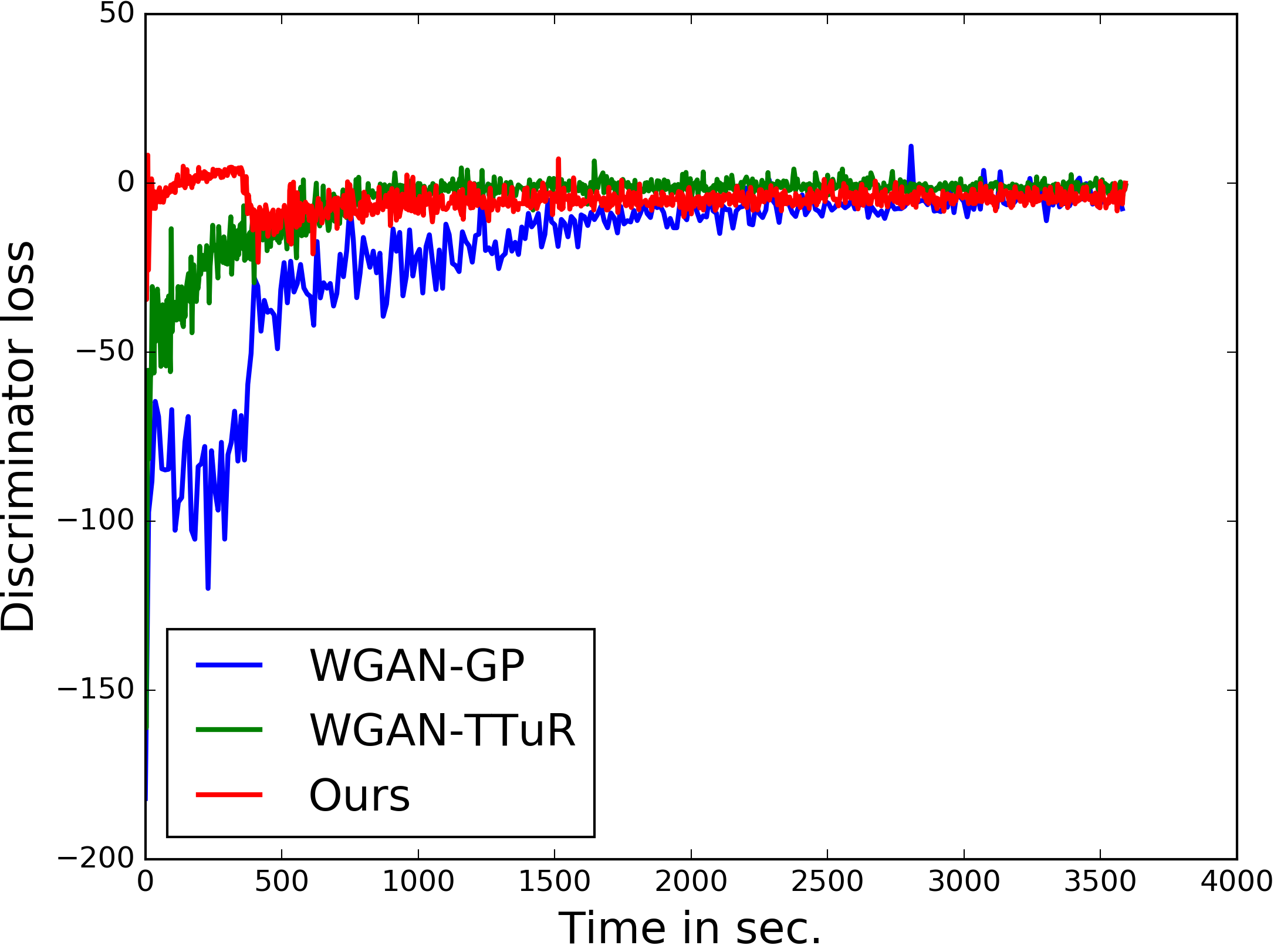} \\
    \caption{Discriminator loss}
  \end{subfigure}
  \begin{subfigure}{0.49\linewidth}
    \includegraphics[width=\linewidth]{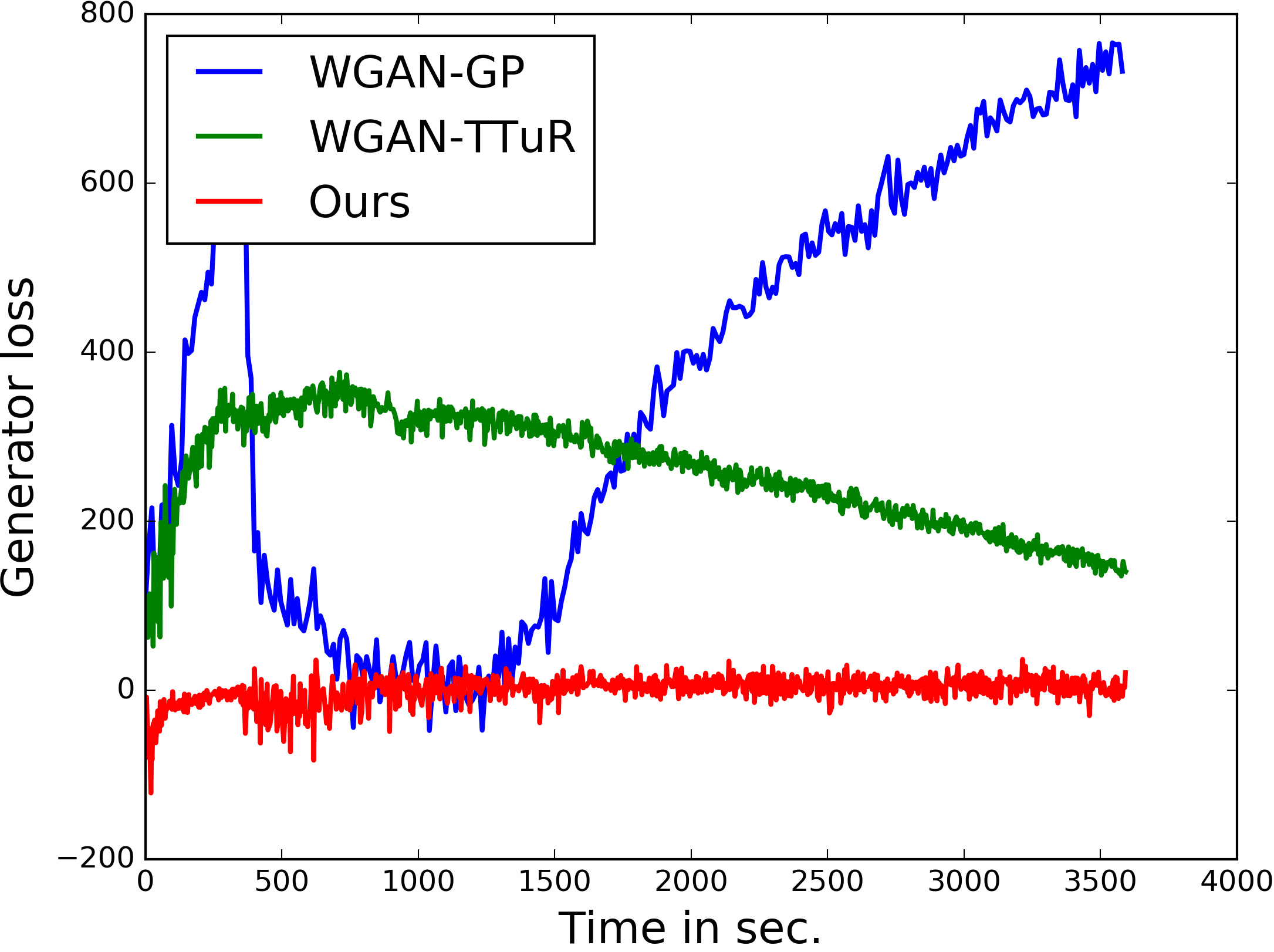} \\
    \caption{Generator loss}
  \end{subfigure}
  \caption{Wall-clock aligned discriminator and generator loss curves, which resulted from training the models on CelebA.}
  \label{fig:celeba_losses}
\end{figure}

\section{Proposed metric to compare generalization capabilities}
In this section, we further elaborate on the design of the generalization score in our proposed metric.
Let $W_{i,j}$ be the Wasserstein distance estimated by discriminator of the $i$-th Wasserstein-GAN between unseen real samples and data, which was generated using the generator from the $j$-th Wasserstein-GAN.
Furthermore, let $W'_{i}$ be the baseline estimate for the $i$-th Wasserstein-GAN, which is computed using its own generator and the trainings dataset.
We define the difference $W_{i,j} - W'_{i}$ as a measure for increase or decrease in generalization capability.
However, to compare the differences between $W_{i',j}$ and $W_{i,j}$, which result from different indices $i$ and $i'$ we have to ensure that the distances have the same scale.
If a discriminator of a Wasserstein-GAN has a Lipschitz constant of $k$ it estimates $k \cdot W_1(\mathbb{P}_r, \mathbb{P}_g)$, which implies that the distance for $i$ and $i'$ could have different scales as well.
To avoid such scaling problems, we define the generalization score as
\begin{equation}
  \begin{split}
  \frac{k \cdot (W_{i,j} - W'_{i})}{| k \cdot W'_{i}|} &= \frac{k \cdot (W_{i,j} - W'_{i})}{k \cdot |W'_{i}|} \\
  &= \frac{W_{i,j} - W'_{i}}{ |W'_{i}|}.
  \end{split}
\end{equation}
where the influence of the positive constant $k$ is cancelled out.
\section{Effect of mode preservation on image quality and loss behaviour}
Figures~\ref{fig:mnist}, ~\ref{fig:cifar10} and ~\ref{fig:celeba} demonstrate the effect of the better mode preservation regarding the resulting image quality of our approach in comparison to WGAN-GP and WGAN-TTUR for different datasets.
For all data sets, WGAN-GP generates samples that show significantly more distortions and artefacts than the other methods.
While WGAN-TTUR is able to create more realistic samples than WGAN-GP, it still generates more artefacts than our approach.
This is especially prevalent in Figures \ref{fig:mnist} and \ref{fig:celeba}.
In addition, we provide the loss characteristics in Figures~\ref{fig:cifar10_losses} and ~\ref{fig:celeba_losses}.
Note that both the generator loss and discriminator loss converge when using the proposed training procedure while the other algorithms can lead to a diverging generator loss.

\end{document}